\documentclass{article}

 \usepackage[main, final]{neurips_2026}

\usepackage[utf8]{inputenc} 
\usepackage[T1]{fontenc}    
\usepackage{hyperref}       
\usepackage{url}            
\usepackage{booktabs}       
\usepackage{amsfonts}       
\usepackage{nicefrac}       
\usepackage{microtype}      
\usepackage{xcolor}         
\usepackage{graphicx}
\usepackage{amsmath}
\usepackage{caption}
\usepackage{subcaption}
\usepackage{amsthm}

\theoremstyle{plain}
\newtheorem{theorem}{Theorem}[section]

\newtheorem{proposition}[theorem]{Proposition}
\newtheorem{remark}[theorem]{Remark}

\title{Supervised Distributional Reduction via Optimal
Transport and Dependence Maximization}
\author{%
Sai-Aakash Ramesh$^{1,2}$\thanks{Correspondence to Sai-Aakash Ramesh: <\texttt{sai-aakash.ramesh@digilab.ai}>} \quad Archit Sood$^{1}$ \quad Andrew Corbett$^{1,2}$ \quad Tim Dodwell$^{1,2}$\\
\\ $^1$digiLab, UK \ $^2$University of Bristol, UK\\
}

\begin{document}

\maketitle

\begin{abstract}
  
    Learning representations that capture both intrinsic data geometry and target-relevant structure remains a fundamental challenge, particularly in settings where data reduction must balance compression with predictive fidelity. While distributional reduction—encompassing joint clustering and dimensionality reduction—offers a principled way to summarize data, its supervised variants remain relatively under-explored, despite the importance of retaining task-relevant signal for downstream prediction and decision-making. We propose Supervised Distributional Reduction (SDR), an algorithm for learning target-aware representations by combining optimal transport with explicit dependence maximization. SDR builds on the Fused Gromov–Wasserstein (FGW) objective to align the relational structure of the input distribution with a set of representative points, while augmenting it with a direct dependence term that encourages the learned embeddings to capture predictive signal more explicitly. This results in compact representations that reflect both geometric structure and supervision. Beyond representation learning, SDR naturally induces a data-dependent, non-stationary geometry that can be leveraged for settings such as Gaussian Process (GP) modelling. By redefining distances through target-aware distributional alignment, SDR enables the construction of adaptive kernels that respond to local variations in both data geometry and supervision, offering an optimal transport-based perspective on non-stationary kernel design.

\end{abstract}

\section{Introduction}
\label{sec:introduction}

    Optimal Transport (OT) provides a principled framework for comparing probability distributions, with Gromov–Wasserstein (GW) \citep{peyre2016gw} enabling alignment across heterogeneous spaces by matching relational structure. Building on this, Distributional Reduction (DistR) \citep{vanassel2025distr} learns a low-dimensional set of representative points that preserves the geometry of a dataset, providing a framework for unifying clustering and dimensionality reduction. However, existing distributional reduction methods are largely unsupervised, which can lead to representations that fail to retain task-relevant information and lack a clear mechanism for out-of-sample generalization. This prevents these methods from being used for downstream prediction tasks.  

    We address these limitations by proposing Supervised Distributional Reduction (SDR), a framework for learning target-aware latent representations by combining optimal transport with explicit dependence maximization. SDR builds on the Fused Gromov–Wasserstein (FGW) objective \citep{vayer2018fgw} to align the relational structure of the input distribution with representative points, while incorporating supervision. Crucially, rather than relying solely on the coupling matrix—where supervision is mediated indirectly and may be diluted by structural constraints—we augment the objective with a direct representation-level dependence term based on Centered Kernel Alignment (CKA). This enables explicit alignment between embeddings and targets while preserving intrinsic data geometry.

    Beyond representation learning, SDR induces a data-dependent, non-stationary geometry that can be exploited for kernel learning. By constraining embeddings to admit an approximate kernel-based out-of-sample extension, we obtain a stable projection mechanism for unseen data. This allows SDR to define adaptive covariance functions for Gaussian Processes (GPs), where the learned geometry translates into non-stationary kernel behaviour that responds to both structure and supervision in the data.

\subsection{Contributions}
\label{sec:contributions}

We make the following contributions in this work:
\begin{enumerate}
    \item \textbf{Supervised distributional reduction via OT and dependence maximization:} We propose SDR, an algorithm that combines OT-based alignment with explicit dependence maximization to learn compact, target-aware representations that preserve both geometry and predictive signal.
    \item \textbf{Out-of-sample extension via RKHS projection with applications to GPs:} We formulate the mapping from inputs to embeddings as a regularized kernel ridge regression problem, yielding an out-of-sample transformation map that allows SDR to be used as part of predictive pipelines and for non-stationary kernel construction in GPs.
\end{enumerate}

\begin{figure}[t]
    \centering
    \includegraphics[width=0.9\linewidth]{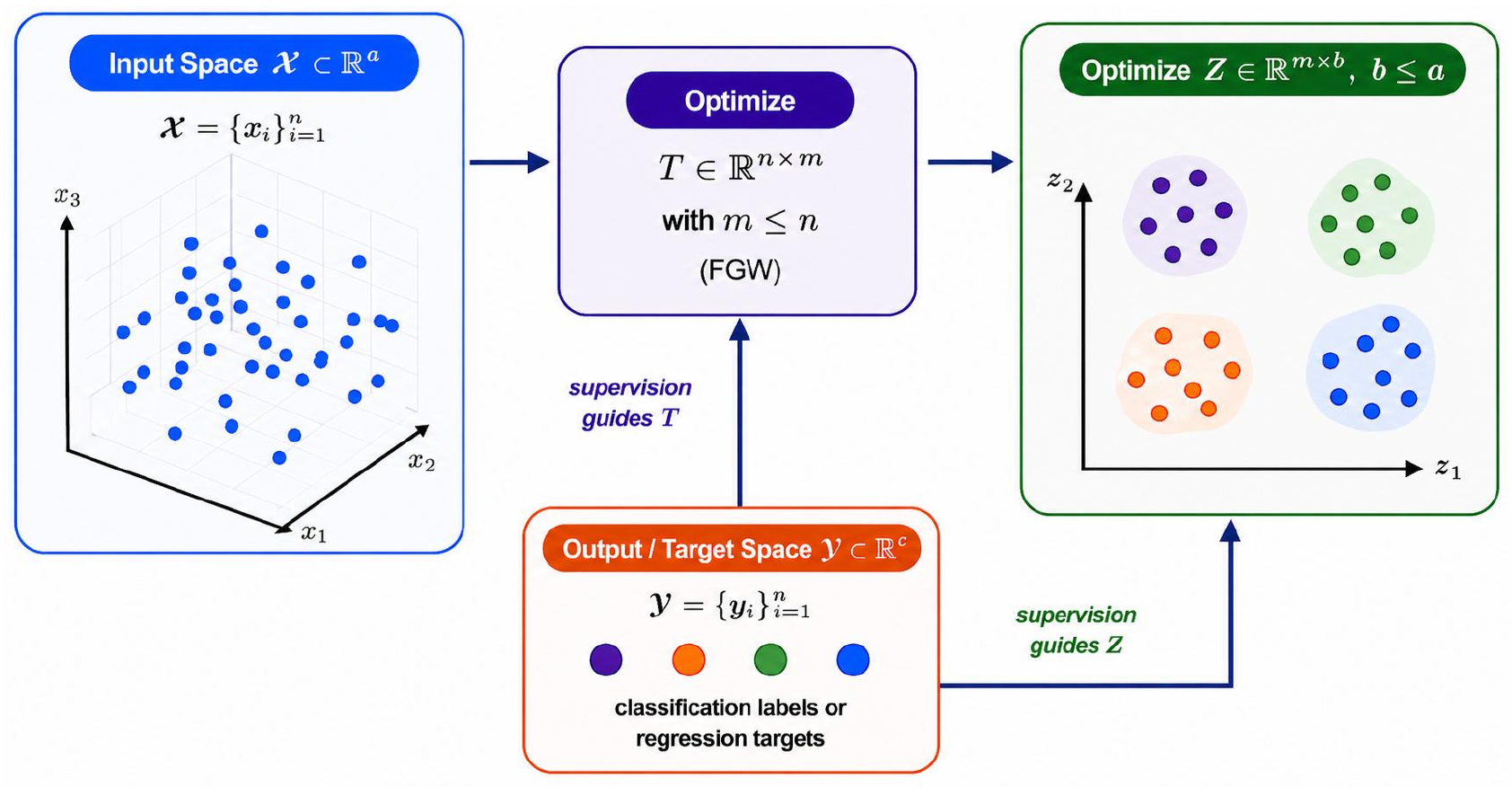}
    \caption{Supervised Distributional Reduction (SDR) with Fused Gromov-Wasserstein OT formulation and explicit dependence maximization for target-aware distributional reduction.}
    \label{fig:sdr}
\end{figure}

\section{Related Work}
\label{sec:related}

\subsection{Supervised Dimensionality Reduction and Clustering}
Supervised dimensionality reduction (DR) methods incorporate label information to learn representations that are more discriminative for downstream tasks. Classical approaches such as Linear Discriminant Analysis (LDA) \citep{fisher1936lda} maximize class separability through linear projections, while metric learning methods, including Neighbourhood Components Analysis (NCA) \citep{goldberger2004nca} and Large Margin Nearest Neighbour (LMNN) \citep{weinberger2005lmnn}, shape the embedding space using supervised distance objectives. Extensions of unsupervised DR techniques have also been proposed in supervised settings; for instance, Supervised UMAP \citep{mcinnes2020umap} integrates label information to guide the preservation of local structure in a target-aware manner. In parallel, dependence-based approaches maximize statistical dependence between representations and targets. For example, using the Hilbert Schmidt Independence Criterion (HSIC) \citep{gretton2007hsic} in conjunction with PCA gives rise to a version of Supervised PCA \citep{barshan2011spca}. On the other hand, supervised and semi-supervised clustering methods introduce constraints such as must-link and cannot-link pairs to guide cluster formation \citep{wagstaff2001constrainedkmeans}. \cite{xing2002clusteringwithsideinfo} suggest providing supervision in the form of distances (similarities) for metric-informed clustering. Despite these advances, most approaches either focus solely on embedding or clustering, or rely on heuristic combinations of the two, lacking a unified formulation that captures both aspects at the distributional level.

Alongside the rise of deep learning came the paradigm of deep metric learning that used neural networks to learn similarity measures between inputs. Deep metric learning introduced target dependent objectives like contrastive, triplet and proxy-based losses(\citep{Schroff_2015_CVPR}, \citep{khosla2020supcon}, \citep{kim2020proxyanchorlossdeep}) that produced models which shape embedding geometry for discriminability alone. Prototype-based methods such as Prototypical Networks (\citep{snell2017protonet}) and Learning Vector Quantization (LVQ) methods like General LVQ (\citep{sato1995glvq}) were introduced for classification problems and typically represent each class using a single prototype, thereby collapsing all within-class variation. SDR instead retains input-space structure through an explicit structure-preserving term. Its prototypes are free variables inferred jointly with the embedding, rather than class means fixed in number and identity by the labels, and carry a mass-conserving transport coupling in place of a softmax over distances. Treating labels as features further admits continuous targets, which these classification methods do not.

\subsection{Optimal Transport, GW/FGW, and Distributional Reduction}

Optimal transport (OT) provides a principled framework for comparing probability distributions by minimizing the cost of transporting probability mass between them \citep{monge1781memoire, kantorovich1958ot}. Beyond classical Wasserstein distances, Gromov–Wasserstein (GW) extends OT to settings where only relational information, such as pairwise distances, is available, enabling comparisons across heterogeneous spaces \citep{peyre2016gw}. The Fused Gromov–Wasserstein (FGW) formulation further combines feature-level and structural alignment by integrating Wasserstein and GW objectives into a single framework \citep{vayer2018fgw}. These formulations provide a natural mechanism for geometry-aware representation learning. Building on these ideas, OT-based approaches have been proposed for distributional summarization, including Wasserstein barycenters \citep{cuturi14wbarycenters} and reduced-support approximations \citep{forrow2019otfactoredcouplings}. Methods such as DistR \citep{vanassel2025distr} leverage GW to jointly perform clustering and dimensionality reduction by learning a reduced set of representative points that preserve dataset geometry. However, these approaches are largely unsupervised and do not explicitly incorporate target information during the reduction process. 

SDR builds on this line of work by integrating supervision directly into the distributional reduction objective, enabling the learning of compact representations that are both geometry-preserving and target-aware.

\section{Method}
\label{sec:Method}

\subsection{Distributional Reduction with OT}
Distributional Reduction (DistR) \citep{vanassel2025distr} formulates joint clustering and DR as an optimal transport problem between the original dataset and a reduced set of representative embeddings. Given input data $X \in \mathbb{R}^{n \times d}$, DistR seeks a reduced representation $Z \in \mathbb{R}^{m \times p}$ with $m \leq n$ and $p \leq d$ by solving a Gromov–Wasserstein (GW) OT problem between their respective metric structures.

Let $C$ and $\bar{C}$ denote pairwise distance matrices in the input and embedding spaces respectively and $\textbf{h}_{X}$ denote the weight vector typically assumed to be uniform, that is $\textbf{h}_{X} = \frac{1}{n}\textbf{1}_{n}$. DistR solves:
\begin{equation*}
    \underset{\substack{\textbf{Z}\in \mathbb{R}^{m \times p}\\ \textbf{T} \in \mathcal{U}(\textbf{h}_{X}) \\
     \textbf{h}_{Z} \in \Sigma_{m}}}{\text{min}} \sum_{ijkl} \ L_{u}(C_{ij}, \bar{C}_{kl})T_{ik}T_{jl} \tag{DistR}
\end{equation*}
where $T$ is a transport plan in $\mathcal{U}(\textbf{h}_{X}) = \{T \in \mathbb{R}_{+}^{n \times m}: T\textbf{1}_{m} = \textbf{h}_{X}\}$ between the input and reduced representations, and $L_{u}$ is an unsupervised, structural loss function comparing pairwise distances. The problem is optimized via alternating minimization over $T$ and $Z$, where $T$ is computed using standard GW solvers and $Z$ is updated using gradient-based methods. When $m < n$, this yields a reduced set of representative points weighted non-uniformly with $h_{Z}$, unifying clustering and dimensionality reduction within a single framework.

\subsection{Supervised Distributional Reduction}

In this section, we extend the Distributional Reduction Algorithm to the supervised setting. SDR optimizes a single unified objective over $(Z, T)$, where alternating minimization is used to handle the non-convex coupling between the variables. The SDR objective solves the following two subproblems in the $T$ and $Z$ update steps:
\begin{itemize}
    \item $T$-step: Solving a Fused-Gromov Wasserstein (FGW) alignment subproblem for $T$
    \item $Z$-step: Optimizing a supervised representation subproblem for $Z$
\end{itemize}

\subsubsection{Optimal Transport with Supervised FGW}
The Fused Gromov-Wasserstein (FGW) \citep{vayer2018fgw} objective is formulated as follows:
\begin{equation}
    FGW_{\alpha}(C, \bar{C}, Y, h, \bar{h}; T) = \underset{T \in \mathcal{U}(h, \bar{h})}{\text{min}} (1 - \alpha) \sum_{ij} L_{s}(y_{i}, g_{j}) T_{ij} + \alpha \sum_{ijkl} \ L_{u}(C_{ij}, \bar{C}_{kl}) T_{ik}T_{jl}
    \label{eq:sfgw}
\end{equation}

where, $y_{i}$ and $g_{j}$ are features of a point $x_{i}$ in the input space and a point $z_{j}$ in the embedding space respectively, $L_{s}$ is a feature-level loss function, and $\alpha$ is a hyperparameter that controls the magnitudes of both these losses. In this work, we specifically choose the target values — regression targets or class labels present in a dataset as the features. This objective is intractable in conventional ML settings because prototype targets $g_j$ are unavailable (in graph settings, by contrast, both graphs in question can have observed node-level labels or features). That makes $\textbf{G} = (g_{1}, g_{2}, ..., g_{m})$, additional parameters to be optimized as part of the objective. For notational convenience, we make the summation variables explicit, abbreviate the GW loss and rewrite the objective in Eqn.\eqref{eq:sfgw} as follows:
\begin{equation*}
    \mathcal{J}(T, Z, G) = (1 - \alpha) \sum_{i=1}^{n}\sum_{j=1}^{m}L_{s}(y_{i}, g_{j})T_{ij} + \alpha GW(Z; T) \tag{S-FGW}
\end{equation*}

The key insight is that we can analytically eliminate the prototype targets $g_{j}$ from the S-FGW objective by solving for a partially minimized objective as described in Remark \ref{rem:partial_min}. Proposition \ref{prop:proto_targets} shows how we can eliminate prototype targets in closed form by leveraging the Bregman barycentric structure, yielding a tractable objective that depends only on $(T, Z)$ for common loss functions. 

\begin{remark}[Elimination of Prototype Targets by Partial Minimization]
    \label{rem:partial_min}
    \leavevmode \\
    The supervised FGW objective is separable in the prototype targets $G = \{g\}_{j=1}^{m}$. For fixed $T$ and $Z$, each $g_{j}$ can therefore be optimized independently as
    \begin{equation}
        g_{j}^{*}(T) = \underset{g \in \mathcal{G} }{\text{arg min}} \ \sum_{i=1}^{n}L_{s}(y_{i}, g)T_{ij}
    \end{equation}
    Substituting these minimizers yield an equivalent reduced objective depending only on $(T, Z)$,
    \begin{equation}
        \underset{T, Z, G}{\text{min}}\mathcal{J}(T, Z, G) = \underset{T, Z}{\text{min}} \mathcal{J}(T, Z, G^{*}(T))
    \end{equation}
\end{remark}

\begin{proposition}[Optimal Prototypes for Bregman Losses; adapted from \cite{banerjee2005bregmancentroids} and \cite{nielson2009symmetrizedcentroids}]
    \label{prop:proto_targets}
    \leavevmode\\
    Let $L_{s}$ be a Bregman divergence generated by a differentiable, strictly convex function $\phi$ on a convex domain $\mathcal{G} \subseteq \mathbb{R}^{c}$ and let $\pi_{j} := \Sigma_{i=1}^{n}T_{ij} > 0$. Then the optimal prototype target satisfies:
    \begin{equation}
        g_{j}^{*}(T) = 
        \begin{cases}
            (\nabla \phi)^{-1} \left( \frac{1}{\pi_{j}}\sum_{i=1}^{n}T_{ij} \nabla \phi(y_{i})\right), \ \textit{\text{($g_{j}$ is first argument in $L_{s}$)}} \\
            \frac{1}{\pi_{j}}\sum_{i=1}^{n}T_{ij}y_{i}, \ \textit{\text{($g_{j}$ is second argument in $L_{s}$)}}
        \end{cases}
    \end{equation}
    \leavevmode\\
    For standard supervised losses used in practice particularly the L2-loss for regression and the cross-entropy loss for classification, the optimal prototype targets reduce to:
    \begin{equation}
        g_{j}^{*}(T) = \frac{\sum_{i=1}^{n}T_{ij}y_{i}}{\sum_{i=1}^{n}T_{ij}} = \frac{1}{\pi_{j}}\sum_{i=1}^{n}T_{ij}y_{i}
    \end{equation}
\end{proposition}

The proof for Proposition \ref{prop:proto_targets} is provided in Appendix \ref{appx:prop1}. Substituting $g_j^*(T)$ yields a reduced objective depending only on $(T,Z)$, which we refer to as the barycentric supervised FGW (BS-FGW) objective.
\begin{equation*} 
    \mathcal{J}(T, Z) = (1 - \alpha) \sum_{i=1}^{n}\sum_{j=1}^{m}L_{s}(y_{i}, g_{j}^{*}(T))T_{ij} + \alpha GW(Z; T) \tag{BSFGW}
\end{equation*}

The BSFGW problem is posed over $\mathcal{U}(\textbf{h}_{X}, \textbf{h}_{Z})$ which fixes both marginals. In our setting, $\textbf{h}_{Z}$ is an unknown quantity. We therefore optimize over the semi-relaxed set $\mathcal{U}(\textbf{h}_{X})$ which constrains only the first marginal as introduced in \citep{vincentcuaz2022srgw}. The second marginal is left free and recovered from the solution as $\textbf{h}_{Z} = \textbf{T}^\top \textbf{1}_{n}$ mirroring the DistR approach. The $\textbf{T}$-subproblem is now defined as:
\begin{equation*}
    \textbf{T}^{*} = \underset{\textbf{T} \in \mathcal{U}(\textbf{h}_{X})}{\text{arg min}} (1 - \alpha) \sum_{i=1}^{n} \sum_{j=1}^{m} L_{s}(y_{i}, g_{j}^{*}(T)) T_{ij} + \alpha GW(Z; T)
    \tag{srBSFGW}
\end{equation*}

\subsubsection{Supervised Representation Optimization}
The DistR objective is elegant in the sense that it uses the same GW loss to optimize for both $Z$ and $T$. However, this is not as straightforward in the supervised setting.
The BSFGW objective introduces supervision indirectly via $T$, and the gradient of the supervision term with respect to $Z$ is inert almost everywhere as described in the following remark.

\begin{remark}
    \label{rem:fgw_bottleneck}
    The partially minimized BSFGW objective exhibits two key limitations:
    \begin{itemize}
        \item [(i)]\textbf{No direct supervision in representation updates - } For fixed $T$, the supervised term does not depend on $Z$, and the representation update is governed solely by the geometric term.

        \item [(ii)]\textbf{Gradient-inertness of supervision for the representation variable holding a.e for the joint objective and for any first-order method - } 
        Let $F(Z) = \underset{T}{\text{min}} \ J(Z, T)$, then the gradient $\nabla F(Z)$ is practically inert wherever $T^*$ is unique and $F$ is differentiable. Supervision influences $Z$ only through discrete switches on the coupling polytope which is not visible to a first-order optimization method.
    \end{itemize}
\end{remark}

 In regimes where $T^*(Z)$ is dominated by structural constraints or is locally insensitive to $Z$, the supervised gradient on $Z$ is attenuated leading to weak alignment between Z and Y. Theoretical justification for Remark \ref{rem:fgw_bottleneck} is provided in Appendix \ref{appx:fgw_bottleneck_theory}.

To address this, we introduce a direct dependence-maximization term, which complements FGW by providing strong gradients on Z. We inject an explicit supervision term using a normalized version of the HSIC score called the \textbf{Centred Kernel Alignment (CKA)} \citep{kornblith2019cka} which is defined over two variables $Z$ and $Y$ as:
\begin{equation*}
    \text{CKA}(Z, Y) = \frac{\langle K_{Z}, K_{Y} \rangle _{\mathcal{F}}}{||K_{Z}||_{\mathcal{F}}||K_{Y}||_{\mathcal{F}}} \tag{CKA}
\end{equation*}
where, $K_{Z}$ and $K_{Y}$ are kernel matrices over $Z$ and $Y$ respectively and $||.||_{\mathcal{F}}$ denotes the Frobenius norm. We use CKA as a dependence maximization objective to directly align representations with targets, building on prior work that employs similar dependence-based criteria (e.g., HSIC) for supervised dimensionality reduction \citep{barshan2011spca, gretton2007hsic, Zhang2017hsic}. Unlike pointwise losses, CKA operates at the representation level and captures global dependence consistent with the global, structure-preserving perspective of OT. The normalized range of CKA makes it interpretable and offers easier tuning for the hyperparameter $\eta$ when combined with the GW loss in the SDR objective. While a GW-style residual on $K_Z, K_{\tilde{Y}}$ seems like a natural option here, this can be counterproductive. For example, with discrete labels, $K_Y$ is block constant and minimizing $||K_{Z} - K_{\tilde{Y}}||^2$ pushes within-class distances toward a constant resulting in class collapse. This destroys the intra-class geometry the GW term preserves. On the other hand, CKA maximizes dependence rather than kernel equality making it invariant to isotropic scaling and orthogonal transforms. This aligns $Z$ with $Y$ while leaving degrees of freedom for structure preservation.

We fix $K_{Z}$ to be the PSD kernel whose decay profile matches that of the embedding affinity $\bar{C}$ (for example $\bar{C} \rightarrow \text{Student-t}\implies K_{Z} \rightarrow \text{RQ}, \bar{C} \rightarrow \text{Gaussian}\implies K_{Z} \rightarrow \text{RBF}$). We chose this convention so CKA and GW gradients act on a consistent geometry making the method transferable to similar affinity-distance pairs. For $m < n$, we compute CKA using a coupling-induced projection $K_{\tilde{Y}\tilde{Y}} = T^{\top}K_{YY}T$, which defines target similarities in the reduced space. This can be viewed as a transport-weighted kernel compression technique that allows CKA computation in the prototype space. In Appendix \ref{appx:fgw_bottleneck_evidence} we provide some empirical evidence on the effect of the additional CKA term compared to the vanilla BSFGW objective. Combining the CKA term with the BSFGW objective, the joint optimization objective for $\textbf{Z}$ and $\textbf{T}$ is given by:
\begin{equation*}
    J_{\text{SDR}}(Z, T, h_{Z}) = 
    (1 - \alpha) \ \sum_{i=1}^{n}\sum_{j=1}^{m} L_{s}(y_{i}, g_{j}^{*}(T)) T_{ij} 
    + \alpha \ GW(Z; T)
    - \eta \ \text{CKA}(Z, \tilde{Y}) \tag{SDR}
\end{equation*} 

By minimizing the negative of the CKA term we maximize the correlation between the variables $Z$ and $\tilde{Y}$ with $\eta$ controlling the relative contribution of the CKA term to the objective. We minimize $J_{\text{SDR}}$ to obtain the embeddings $Z$, coupling matrix $T$ and the weighting $h_{Z}$ of the embeddings:
\begin{equation}
    \textbf{Z}^{*}, \textbf{T}^{*}, \textbf{h}_{Z}^{*} = \underset{\substack{\textbf{Z}\in \mathbb{R}^{m \times p} \\
        \textbf{T} \in \mathcal{U}(\textbf{h}_{X}) \\
        \textbf{h}_{Z} \in \Sigma_{m}}}{\text{arg min}} \ J_{\text{SDR}}(Z, T, h_{Z})
\end{equation}
We adopt an inexact alternating scheme w.r.t $T$ to solve for the SDR objective. In particular, the T-step optimizes the srBSFGW objective ignoring the CKA term. This design is intentional for two reasons: (i) including the CKA term in the transport subproblem would break the structure required for efficient OT solvers and introduce high-order, normalized dependencies on T; and (ii) CKA is designed to provide direct supervision on the representation Z, addressing the gradient inertness of the supervision term without requiring gradients through the transport solver. We show that the worst-case mismatch of a $T$-update is bounded by the user-chosen parameter $\eta$ in Proposition \ref{prop:eta_bound}.
\begin{proposition}[Upper bound on worst-case mismatch of T-update]
    \label{prop:eta_bound}
    Optimizing the srBSFGW objective instead of the complete SDR objective results in a worst-case mismatch for the T-update at most by a value of $\eta$.
    \begin{equation}
        J(T^*) \leq J(T) + \eta
    \end{equation}
\end{proposition}
\begin{proof}
    Let,
    \begin{equation}
        J(T) = F(T) - \eta C(T)
    \end{equation}
    where $F$ is the srBSFGW objective and $C(T) = \text{CKA}(Z, Y(T)) \in [0, 1]$. The $T$-step returns $T^* \in \underset{T}{\text{arg min}} \ F$, hence for any feasible T:
    \begin{align}
        J(T^*) & = F(T^*) - \eta C(T^*) \\
        & \leq F(T) - \eta C(T^*) \\
        & = J(T) + \eta(C(T) - C(T^*)) \\
        & \leq J(T) + \eta
    \end{align}
\end{proof}
Each $T$-step thus decreases the complete SDR objective up to an additive error uniformly bounded by the user-chosen hyperparameter $\eta$, and the $Z$-step is standard descent on the full objective. The srBSFGW objective is solved using a CG solver (\cite{jaggi2013cg}, \cite{vincentcuaz2022srgw}) to optimize for $T$ which will consequently recover $\textbf{h}_{Z} = T^{\top}\textbf{1}_{n}$ as the second marginal. The $Z$-step optimizes the sum of the GW and CKA terms using SGD optimizers like Adam \citep{kingma2017adam}. We also implemented a safeguarded variant of this algorithm with an accept-reject step gated on the SDR objective which accepts a T-update only if it does not increase the full SDR objective. The results for this variant are shown in Appendix \ref{appx:safeguard}. Despite being more stable, such a variant tends to perform empirically worse than the inexact alternating scheme since, a reject-only variant can be stuck in a rejection loop. It is worth noting that SDR inherits the same runtime complexity as DistR which is $\mathcal{O}(nm^{2} + mn^{2})$ according to Proposition 1 in \citep{peyre2016gw}.

\section{Mapping Estimation for SDR}
\label{sec:oos-extension}
While SDR learns embeddings that are strongly aligned with targets, these representations are not guaranteed to admit a functional mapping from the input space. This lack of an explicit mapping prevents SDR from being used in downstream predictive settings, where embeddings for unseen inputs must be computed. To enable out-of-sample generalization, we therefore seek embeddings $Z$ that are not only target-aware but also approximately representable as a kernel map from the inputs. Concretely, we enforce that $Z$ lies close to the image of a function in the RKHS induced by a kernel $K$, leading to the approximation $Z \approx KL$, where $L$ parametrizes the mapping. This perspective allows us to learn embeddings that retain the advantages of SDR while ensuring they can be consistently extended to unseen data. Consequently, this allows SDR to be used as part of a predictive pipeline.

We take inspiration from \cite{perrot2016map} to propose a modified objective which augments the original SDR objective with a projection-consistency term as follows
\begin{equation*}
    J_{\text{SDR-OOS}}(Z, T, L) =  J_{\text{SDR}}(Z, T)  + \frac{\mu}{2}|| TZ - KL||_{F}^{2} + \frac{\lambda_{L}}{2}\text{Tr}(L^{\top}KL) \tag{SDR-OOS}
\end{equation*}
where, $K \in \mathbb{R}^{n \times n}$ is the covariance matrix of the training points $X$ according to a kernel function (typically RBF), $L \in \mathbb{R}^{n \times p} $ is the transformation map, and $\lambda_{L}$ is a regularization parameter for the transformation map $L$ with the last term defining the RKHS norm of $L$. $\mu$ is a regularization parameter for $Z$ that forces $Z$ to lie close to the RKHS representable set of  the kernel function used to compute $K$. The solution for the L-subproblem can be obtained in closed form and is given by Proposition \ref{prop:krr_sol}. 

\begin{proposition}[Closed form solution for L-subproblem]
    \label{prop:krr_sol}
    The unique minimizer of the L-subproblem in $J_{\text{SDR-OOS}}(\cdot, \cdot, L)$ is given by
    \begin{equation}
        L^{*} = (\mu K + \lambda_{L} I)^{-1} \mu TZ
    \end{equation}
\end{proposition}

The proof is provided in Appendix \ref{appx:l_closed_form} and is a simple least-squares derivation. The SDR-OOS objective can be similarly solved with BCD by having three alternating updates for $T, Z$ and $L$. After updating $L$, we perform a relaxation step that moves $Z$ towards the RKHS-representable set defined by the current projection $KL^{*}$, while also retaining flexibility introduced by the SDR objective. This update can be written down as follows:
\begin{equation}
    Z' = (1 - \beta)Z + \beta KL^{*}
\end{equation}
where $\beta$ controls the trade-off between preserving target-aligned structure in the learned embeddings and enforcing consistency with the RKHS projection for improved generalization. The soft update can be extended to the $m \neq n$ (for SDR typically $m < n$) case analogously as, 
\begin{equation}
    Z' = (I + \beta T^{\text{T}}T)^{-1}(Z + \beta T^{\text{T}}{K}L^{*})
\end{equation}

\subsection{SDR for Non-Stationary Kernel Construction in GPs}
Gaussian Processes (GPs) with stationary kernels are limited in their ability to capture input-dependent variations in function behaviour. A common approach to address this is Deep Kernel Learning (DKL) \citep{wilson16dkl}, where a neural network is jointly trained with the GP to learn a feature representation that induces non-stationarity. While effective, this approach relies on parametric feature extractors and joint optimization of the network parameters and the GP hyperparameters, which can be sensitive to initialization and challenging to train in low-data regimes. Deep Gaussian Processes (DGPs) \citep{damianou2013deepgps} provide another way to introduce non-stationarity by modelling each intermediate representation in a deep belief network with a data-driven GP prior but are expensive to train and converge much slower in practice.

SDR provides an alternative mechanism for constructing non-stationary kernels by learning target-aware representations through optimal transport and dependence maximization in the pure DR setting ($p < d, m = n$). When a stationary kernel (e.g., RBF) is applied in the SDR embedding space, the induced kernel in the original input space becomes data-dependent and non-stationary. The GP prior with an SDR induced kernel can be written as,
\begin{equation*}
    f \sim \mathcal{GP}(m(\cdot), \psi(z(\cdot), z(\cdot))) \tag{SDR-GP}
\end{equation*}
where, $m$ is a mean function, $\psi$ is a stationary kernel like the RBF. The mapping $z(\cdot)$ uses the projection map $L$ optimized by the SDR-OOS objective to project an unseen point $x^*$ and is given by,
\begin{equation}
    z(x^{*}) = K(x^{*}, X)L
\end{equation}
where $K$ is also a stationary kernel like RBF acting in the input space. \

Unlike DKL, this approach decouples representation learning from GP training and leverages a target-aware objective that explicitly aligns representations with targets. The SDR-OOS objective equipped with a projection map allows SDR to serve as a plug-in feature extractor for GPs, providing a non-parametric route to non-stationary kernel design grounded in distributional alignment.

\section{Experiments}
\label{sec:expts}
\subsection{Distributional Reduction Benchmarks}
\label{subsec:distr_benchmarks}
To assess the effect of incorporating supervision, we follow the evaluation protocol used in prior joint clustering and dimensionality reduction methods, particularly DistR. We use three classification datasets namely: COIL-20 \citep{Nene1996coil20}, Fashion-MNIST \citep{xiao2017fmnist} and SNAREseq \citep{chen2019snareseq} for the experiments in this subsection. Further details of the datasets are provided in Appendix \ref{appx:distr_expts}. Since label information Y is explicitly used during optimization, these metrics do not reflect unsupervised clustering performance, but rather the extent to which the learned representation captures target-relevant structure. Accordingly, we report homogeneity score, k-means normalized mutual information (NMI), and silhouette score, interpreting them as measures of label consistency and semantic coherence of the embeddings.

\begin{figure}[h!]
    \centering
    \includegraphics[width=0.95\linewidth]{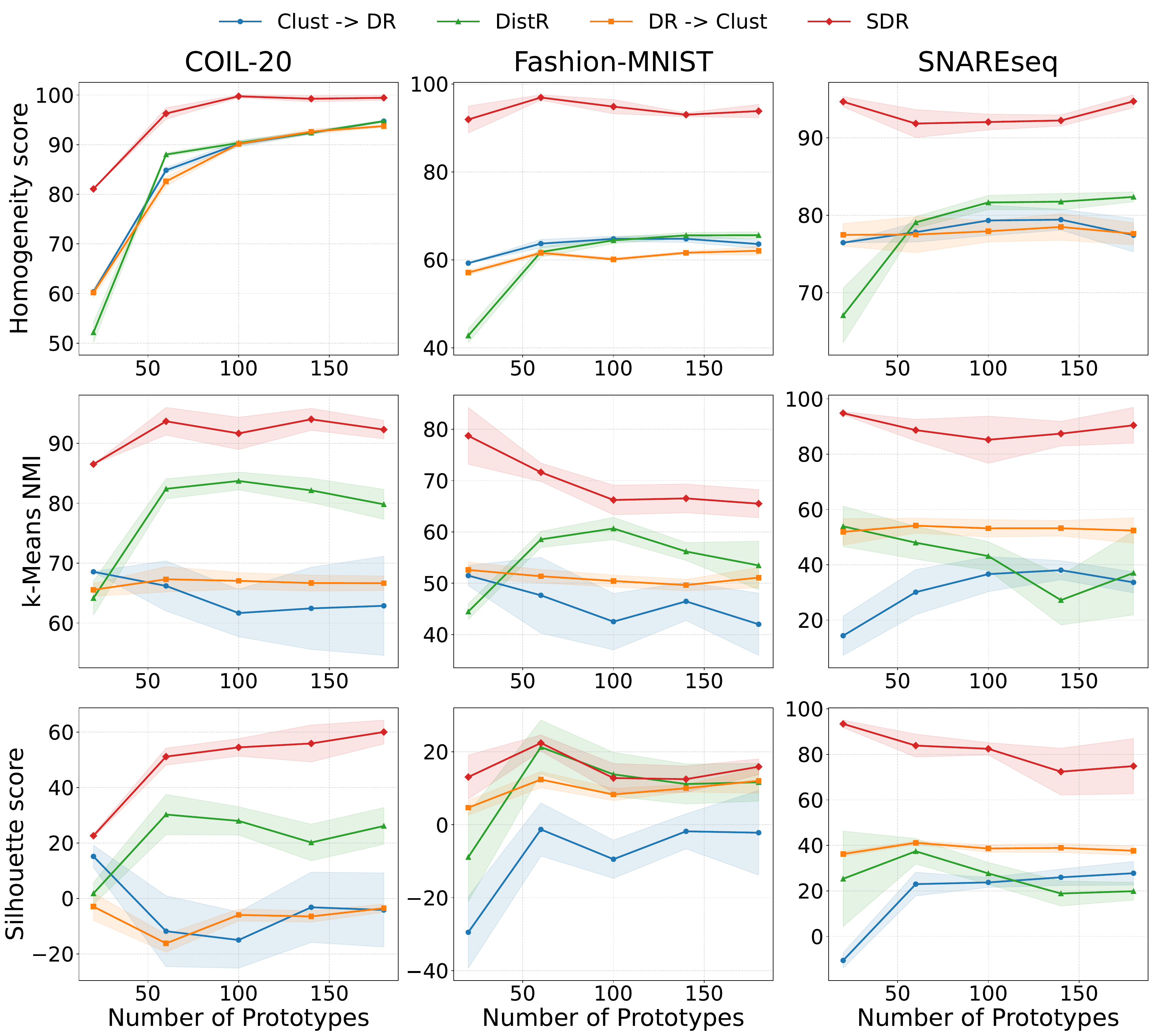}
    \caption{Scores (x100) across varying number of prototypes in $\mathbb{R}^{2}$ for all 4 methods.} 
    \label{fig:joint_clust_dr_metrics}
\end{figure}

\begin{figure}[h!]
    \centering
    \includegraphics[width=1.0\linewidth]{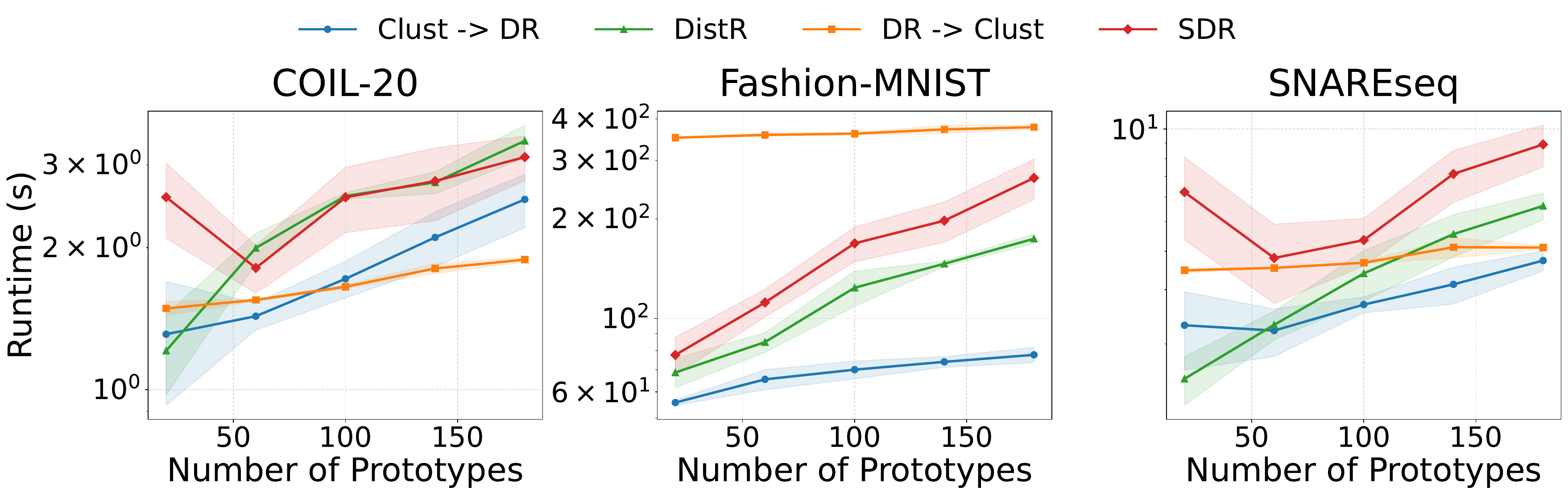}
    \caption{Runtimes across varying number of prototypes in $\mathbb{R}^{2}$ for all 4 methods.} 
    \label{fig:runtimes}
\end{figure}

We compare SDR against DistR, Cluster-then-DR, and DR-then-Cluster to isolate the effect of incorporating supervision within a distributional reduction pipeline. The results of this benchmark are shown in Figure \ref{fig:joint_clust_dr_metrics}, with runtimes plotted in Figure \ref{fig:runtimes}. SDR achieves comparable runtime to DistR with a modest computational overhead. Additional experimental details are provided in Appendix \ref{appx:distr_expts}. An ablation study on the $\eta$ hyperparameter has been performed and the results are reported in Appendix \ref{appx:eta_ablation}. Since SDR jointly reduces the number of samples, predictive evaluation on held-out data is not directly applicable. Instead, we focus on representation-level properties in these experiments, with predictive performance evaluated separately in Section \ref{subsec:kl_expts} under the context of kernel learning when SDR reduces to a pure DR method.

\subsection{Kernel Learning Benchmarks}
\label{subsec:kl_expts}
In this set of experiments, we try to evaluate the effectiveness of SDR as a representation learning method in the context of non-stationary kernel design for GPs. We evaluate on small to medium-sized datasets ($n \leq 2000$) where training flexible, non-stationary GP models can be challenging from an optimization perspective. We focus on this regime where GPs remain practical models, since predictive posterior computation and marginal likelihood evaluation scales as $\mathcal{O}(N^{3})$. SDR exhibits the same $\mathcal{O}(N^{3})$ scaling in the pure DR setting inheriting this computational complexity directly from the DistR framework. 

We test SDR-GP against supervised pure DR methods like neighbour Components Analysis \citep{goldberger2004nca}, Supervised UMAP \citep{mcinnes2020umap} and Kernel Supervised PCA \citep{barshan2011spca} (dubbed as NCA-GP, KSPCA-GP and UMAP-GP respectively). We also compare SDR-GP with DKL and DGP models that are typically considered for non-stationary kernel construction in the GP literature. We evaluate these models across three standard UCI benchmark datasets for regression: Boston Housing, Energy Efficiency and Concrete Compressive Strength \citep{Dua2019UCI} along with two image classification benchmark datasets: MNIST \citep{lecun2010mnist} and COIL-20 \citep{Nene1996coil20}. The Mean Log Likelihood (MLL) and the Mean Squared Error (MSE) for the regression tasks are presented in Table \ref{tab:dkl_regression}. The Mean Log Probability (MLP) for the correct class and Accuracy (ACC) for the classification tasks are reported in Table \ref{tab:dkl_classification}. We also test on a variant of DKL that uses pre-trained feature extractor networks in Appendix \ref{appx:pretrained_dkl}.

\begin{table}[h!]
\captionsetup{skip=12pt}
\centering
\resizebox{\textwidth}{!}{
\begin{tabular}{lcc cc cc}
\toprule
 & \multicolumn{2}{c}{Boston} & \multicolumn{2}{c}{Energy} & \multicolumn{2}{c}{Concrete} \\
\cmidrule(lr){2-3} \cmidrule(lr){4-5} \cmidrule(lr){6-7}
 & MLL ($\uparrow$) & MSE ($\downarrow$) 
 & MLL ($\uparrow$) & MSE ($\downarrow$) 
 & MLL ($\uparrow$) & MSE ($\downarrow$) \\
\midrule
NCA-GP & -0.76 $\pm$ 0.31 & 0.25 $\pm$ 0.11 & -0.81 $\pm$ 0.38 & 0.29 $\pm$ 0.38 & -1.01 $\pm$ 0.16 & 0.45 $\pm$ 0.12 \\
KSPCA-GP & -0.99 $\pm$ 0.27 & 0.38 $\pm$ 0.13 & -0.67 $\pm$ 0.05 & 0.22 $\pm$ 0.03 & -1.28 $\pm$ 0.16 & 0.79 $\pm$ 0.23 \\
UMAP-GP & -1.32 $\pm$ 0.43 & 0.39 $\pm$ 0.11 & -0.53 $\pm$ 0.08 & 0.13 $\pm$ 0.06 & -1.38 $\pm$ 0.19 & 0.67 $\pm$ 0.18 \\
DGP & -1.18 $\pm$ 0.09 & 0.46 $\pm$ 0.18 & -0.88 $\pm$ 0.04 & 0.12 $\pm$ 0.01 & -1.10 $\pm$ 0.05 & 0.33 $\pm$ 0.06 \\
DKL & -0.42 $\pm$ 0.21 & 0.13 $\pm$ 0.07 & -0.45 $\pm$ 0.03 & 0.09 $\pm$ 0.02 & -0.65 $\pm$ 0.11 & 0.21 $\pm$ 0.05 \\
SDR-GP (ours) & -0.32 $\pm$ 0.18 & 0.14 $\pm$ 0.08 & \textbf{-0.37} $\pm$ \textbf{0.02} & \textbf{0.05} $\pm$ \textbf{0.01} & \textbf{-0.40} $\pm$ \textbf{0.07} & \textbf{0.13} $\pm$ \textbf{0.02} \\
\bottomrule
\end{tabular}}
\caption{Test results on UCI regression datasets based on five random seeds}
\label{tab:dkl_regression}
\end{table}

\begin{table}[h!]
\captionsetup{skip=12pt}
\centering
\small
{\setlength{\tabcolsep}{4pt}
\begin{tabular}{lcccc}
\toprule
 & \multicolumn{2}{c}{MNIST-2k} & \multicolumn{2}{c}{COIL-20} \\
\cmidrule(lr){2-3} \cmidrule(lr){4-5}
 & MLP ($\uparrow$) & ACC ($\uparrow$) 
 & MLP ($\uparrow$) & ACC ($\uparrow$) \\
\midrule
NCA-GP & -0.31 $\pm$ 0.04 & 0.91 $\pm$ 0.02 & -0.15 $\pm$ 0.02 & 0.98 $\pm$ 0.01 \\
KSPCA-GP & -0.46 $\pm$ 0.05 & 0.88 $\pm$ 0.02 & -0.13 $\pm$ 0.02 & 0.97 $\pm$ 0.01 \\
UMAP-GP & -1.05 $\pm$ 0.06 & 0.82 $\pm$ 0.00 & -0.90 $\pm$ 0.22 & 0.83 $\pm$ 0.04 \\
DKL & -0.30 $\pm$ 0.05 & 0.90 $\pm$ 0.02 & -0.09 $\pm$ 0.02 & 0.99 $\pm$ 0.00 \\
SDR-GP (ours) & \textbf{-0.23} $\pm$ \textbf{0.02} & \textbf{0.95} $\pm$ \textbf{0.00} & \textbf{-0.07} $\pm$ \textbf{0.02} & 0.99 $\pm$ 0.00 \\
\bottomrule
\end{tabular}
}
\caption{Test results on image datasets based on five random seeds.}
\label{tab:dkl_classification}
\end{table}

\begin{figure}[h!]
    \centering
    \includegraphics[width=0.95\linewidth]{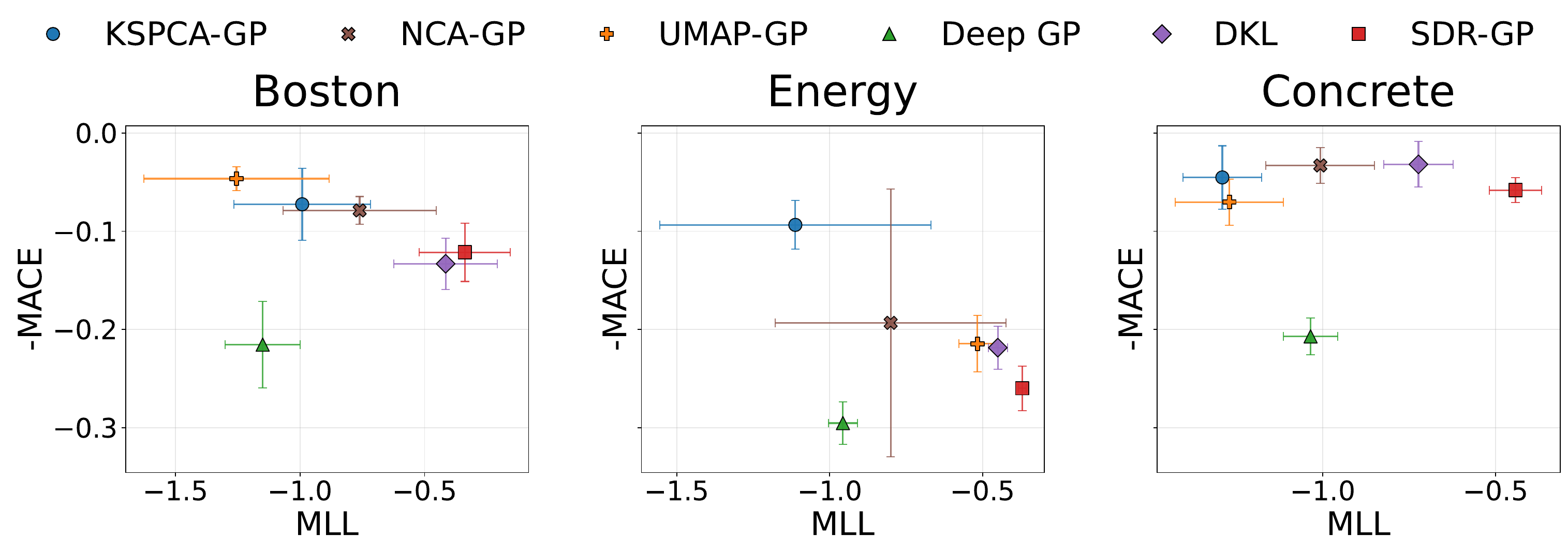}
    \caption{Visualization of different methods depicting the tradeoff between predictive performance in terms of Mean Log Likelihood (MLL) and uncertainty calibration in terms of Mean Absolute Calibration Error (MACE) for the three regression datasets. An ideal model would typically exist in the top-right of this plot which signifies good predictive performance and good uncertainty calibration.}
    \label{fig:mll_mace}
\end{figure}

We evaluate the models not only in terms of predictive performance but also in terms of uncertainty calibration, a key property that makes GPs attractive. Figure \ref{fig:mll_mace} shows how the various models behave in terms of predictive performance and uncertainty calibration. While the SDR objective is aimed at improving predictive performance, a GP trained on SDR representations also provides reasonably calibrated uncertainties (more details in Appendix \ref{appx:calibration}). The details of the datasets, models and hyperparameters used in these experiments are presented in Appendix \ref{appx:kl_expts}. An ablation study on the $\beta$ hyperparameter was performed and the results are reported in Appendix \ref{appx:beta_ablation}.

\section{Conclusion and Future Work}
\label{sec:conclusion}
We introduced Supervised Distributional Reduction (SDR), a framework that integrates optimal transport with dependence maximization to learn compact, target-aware representations. By identifying a gradient inertness problem in the supervision term of the FGW formulation and addressing it through a representation-level objective, SDR provides a mechanism for stronger alignment between embeddings and targets while preserving intrinsic data geometry. We further proposed an out-of-sample extension via RKHS projection that provides an explicit transformation map to project unseen points to the embedding space. This allows SDR to be used in predictive pipelines while also regularizing the optimization of the embeddings. Empirical results demonstrate that SDR improves representation quality and yields strong downstream performance, highlighting the effectiveness of combining transport-based structure with explicit supervision. 

Despite these advancements, we identify a few key areas for improvement. The current version of the algorithm does not provide any convergence guarantees from an optimization theory perspective. Future work can explore optimization alternatives like KL-Mirror Descent for SDR optimization and establish convergence guarantees for the algorithm. Another useful avenue could be to employ low-rank factorizations \citep{scetbon2023lineartimegw, vanassel2025distr} to scale SDR to larger datasets. Exploring alternative dependence measures \citep{Zhang2017hsic} and developing more expressive projection operators for the out-of-sample extension can improve predictive alignment further.

\newpage
\bibliographystyle{plainnat}
\bibliography{refs}


\appendix
\newpage

\section{Proof for Proposition 2 - Optimal Prototype Targets for Bregman Losses}
\label{appx:prop1}
We restate this classical result from Theorem 3.2 in \cite{nielson2009symmetrizedcentroids} for the left-sided Bregman centroid (dual form) and Proposition 1 in \cite{banerjee2005bregmancentroids} for the right-sided Bregman centroids (primal form). We provide a self-contained proof in our notation, since the specific form involving the transport coupling $T$ is needed for the elimination argument in Remark \ref{rem:partial_min} and the BSFGW objective. \\

\textbf{Case 1 ($g_{j}$ as first argument - dual form):}
The definition of a Bregman divergence is given by,
\begin{equation}
    D_{\phi}(g, y_{i}) = \phi(g) - \phi(y_{i}) - \langle \nabla\phi(y_{i}), g - y_{i}\rangle 
\end{equation}

Consider the supervision loss term,
\begin{equation}
    F(g) := \sum_{i=1}^{n}w_{i}D_{\phi}(g, y_{i})
\end{equation}

Expanding the loss term we get,
\begin{equation}
    F(g) = \sum_{i=1}^{n}w_{i}\phi(g) - \sum_{i=1}^{n}w_{i}\phi(y_{i}) - \sum_{i=1}^{n}w_{i} \langle\nabla\phi(y_{i}), g - y_{i} \rangle
\end{equation}

Setting $g = g_{j}$, $w_{i} = T_{ij}$,  $\sum_{i=1}^{n}T_{ij} = \pi_{j}$ we have,
\begin{equation}
    F(g_{j}) = \pi_{j}\phi(g_{j}) - \sum_{i=1}^{n}T_{ij}\phi(y_{i}) - \langle \sum_{i=1}^{n}T_{ij} \nabla\phi(y_{i}), g_{j} \rangle + \sum_{i=1}^{n}T_{ij} \langle \nabla\phi(y_{i}), y_{i} \rangle
\end{equation}

Grouping the terms that only depend on $g_{j}$ we get,
\begin{equation}
    F(g_{j}) = \pi_{j}\phi(g_{j}) - \langle \sum_{i=1}^{n}T_{ij} \nabla\phi(y_{i}), g_{j} \rangle + C
\end{equation}
where $C$ is constant in $g_{j}$. Hence minimizing $F(g)$ is equivalent to minimizing
\begin{equation}
    G(g_{j}) = \pi_{j}\phi(g_{j}) - \langle \sum_{i=1}^{n}T_{ij} \nabla\phi(y_{i}), g_{j} \rangle
\end{equation}

Since $\phi$ is differentiable and strictly convex, $G$ is also differentiable and strictly convex and its unique minimizer is characterized by the first-order condition
\begin{equation}
    \nabla G(g_{j}) = \pi_{j} \nabla \phi (g_{j}) - \sum_{i=1}^{n}T_{ij} \nabla \phi (y_{i}) = 0
\end{equation}

Therefore,
\begin{equation}
    \nabla \phi(g_{j}^{*}) = \frac{1}{\pi_{j}}\sum_{i=1}^{n}T_{ij} \nabla \phi(y_{i})
\end{equation}

Applying $(\nabla\phi)^{-1}$ both sides yields,
\begin{equation}
       g_{j}^{*} = (\nabla \phi)^{-1} \left( \frac{1}{\pi_{j}}\sum_{i=1}^{n}T_{ij} \nabla \phi(y_{i})\right)
\end{equation}

\textbf{Case 2 ($g_{j}$ as second argument - primal form):}
Consider the objective
\begin{equation}
    F(g) := \sum_{i=1}^{n}w_{i}D_{\phi}(y_{i}, g)
\end{equation}

Expanding the objective we get,
\begin{equation}
    F(g) = \sum_{i=1}^{n}w_{i}\phi(y_{i}) - \sum_{i=1}^{n}w_{i}\phi(g) - \sum_{i=1}^{n}w_{i} \langle\nabla\phi(g), y_{i} - g \rangle
\end{equation}

Setting $g = g_{j}$, $w_{i} = T_{ij}$,  $\sum_{i=1}^{n}T_{ij} = \pi_{j}$ we have,
\begin{equation}
    F(g_{j}) = \sum_{i=1}^{n}T_{ij}\phi(y_{i}) - \pi_{j}\phi(g_{j}) - \langle \nabla\phi(g_{j}), \sum_{i=1}^{n}T_{ij}y_{i} - \pi_{j}g_{j} \rangle
\end{equation}

Using the definition of $\bar{y}$ (weighted average of the targets), given by $\Sigma_{i=1}^{n}T_{ij}y_{i} = \pi_{j}\bar{y}$, we get
\begin{align}
    F(g_{j}) & = \sum_{i=1}^{n}T_{ij}\phi(y_{i}) - \pi_{j}\phi(g_{j}) - \pi_{j} \langle \nabla\phi(g_{j}), \bar{y} - g \rangle \\
    & = \sum_{i=1}^{n}T_{ij}\phi(y_{i}) + \pi_{j}(-\phi(g_{j}) - \langle \nabla\phi(g_{j}), \bar{y} - g_{j} \rangle) \\
    & = \sum_{i=1}^{n}T_{ij}\phi(y_{i}) - \pi_{j}\phi(\bar{y}) + \pi_{j}(\phi(\bar{y}) - \phi(g_{j}) - \langle \nabla\phi(g_{j}), \bar{y} - g_{j} \rangle)
\end{align}

Recognizing the Bregman Divergence in the last term
\begin{equation}
    F(g_{j}) = \sum_{i=1}^{n}T_{ij}\phi(y_{i}) - \pi_{j}\phi(\bar{y}) + \pi_{j}D_{\phi}(\bar{y}, g_{j})
\end{equation}

The first two terms do not depend on $g_{j}$. Therefore minimizing $F(g_{j})$ over $g_{j}$ is equivalent to minimizing $D_{\phi}(\bar{y}, g_{j})$. Since $D_{\phi}(\bar{y}, g_{j}) \geq 0$ for all $g \in \mathcal{G}$, with equality iff $g = \bar{y}$ (by strict convexity of $\phi$), the unique minimizer is
\begin{equation}
    g_{j}^{*} = \bar{y}
\end{equation}

Hence, the optimal prototype targets are given by the expression
\begin{equation}
    g_{j}^{*} = \frac{1}{\pi_{j}}\sum_{i=1}^{n}T_{ij}y_{i}
\end{equation}

Now, we derive the closed form expression of the optimal prototype targets for two commonly used loss functions namely: the L2-loss and the Cross-Entropy loss.

\subsection{$L_{s} \rightarrow$ L2-loss}
We can make use of Lemma 2 to derive the optimal prototype targets when $L_{s}$ is the L2-loss. The L2-loss is symmetric w.r.t to the targets and prototype targets and so it does not really make a difference on which case of Lemma 2 we employ to arrive at the expression for $g_{j}^{*}$. In this case, we will just use Case 1 of Lemma 2 assuming $g_{j}$ is the first argument of the $L_{s}$ loss function. For L2-loss we have 
\begin{equation}
    \phi(y) = \frac{1}{2}\left\|y\right\|_{2}^{2}
\end{equation}
The gradient of $\phi$ is then given by, 
\begin{align}
    \nabla \phi(y) & = y \\
    \Rightarrow (\nabla \phi)^{-1}(y) & = y
\end{align}
Lemma 2 then gives,
\begin{align}
    g_{j}^{*}(T) & = (\nabla \phi)^{-1} \left( \frac{1}{\pi_{j}}\sum_{i=1}^{n}T_{ij} \nabla \phi(y_{i})\right) \\
    & = \frac{1}{\pi_{j}}\sum_{i=1}^{n}T_{ij}y_{i}
\end{align}

This just means that the prototype targets are nothing but a weighted sum of all targets weighted according to the coupling matrix $T$.

\subsection{$L_{s} \rightarrow$ Cross-Entropy Loss}
The modified Cross-Entropy (CE) loss as described in \cite{Heskes2025BiasvarianceDT} is given by
\begin{equation}
    L_{s}(y,g) = \sum_{k=1}^{c}y_{k} \text{log}\left(\frac{y_{k}}{g_{k}}\right) + \sum_{k=1}^{c}g_{k} - \sum_{k=1}^{c}y_{k} \tag{Modified CE-Loss}
\end{equation}

The modified CE loss is a valid Bregman divergence with the following generator function
\begin{equation}
    \phi(u) = \sum_{k=1}^{c} u_{k} \text{log}u_{k} - \sum_{k=1}^{c}u_{k}
\end{equation}

The modified CE loss takes $g$ as the second argument and thus Case 2 of Lemma 2 applies. As a direct consequence, the optimal prototype targets are given by
\begin{equation}
    g_{j}^{*} = \frac{1}{\pi_{j}}\sum_{i=1}^{n}T_{ij}y_{i}
\end{equation}

This means that the prototypes in the embedding space carry soft class labels unlike the points in the input space which might have hard labels when one-hot encoded. 

\textbf{Note:} The result established above is general and applies when the domain of $g$ is extended from the probability simplex $\Delta^{c-1}$ to the unit cube $[0, 1]^{c}$ i.e even when we ignore the simplex normalization constraint $\Sigma_{k=1}^{c}g_{k} = 1$. In the case where $y$ is a (one-hot encoded) label vector and $g$ satisfies the simplex normalization constraint, the modified CE loss collapses to the classic CE loss.

\section{Closed form solution for the L-subproblem}
\label{appx:l_closed_form}
The L-subproblem is given by,
\begin{equation}
    J_{\text{SDR-OOS}}(\cdot, \cdot, L) = \psi(L) = \frac{\mu}{2} ||TZ - KL||_{F}^{2} + \frac{\lambda_{L}}{2}\text{Tr}(L^{\top}KL)
\end{equation}
Note that the $L$-subproblem is a quadratic function in $L$. The $L$-subproblem therefore is the sum of a least-squares term with an RKHS norm-regularizer. Since $\lambda_{L} > 0$, the Hessian with respect to $L$ is positive definite. Hence, the $L$-subproblem is strongly convex and admits a unique minimizer. Differentiating $\psi(L)$ w.r.t $L$ and using the first-order optimality condition we have,
\begin{equation}
    \nabla_{L} \psi(L) = \mu K(KL - TZ) + \lambda_{L}KL = 0
\end{equation}
With some basic algebraic manipulations, the minimizer is given by,
\begin{equation}
    L^* = (\mu K + \lambda_{L}I)^{-1}\mu TZ
\end{equation}

Estimating the OOS map reduces to vector-valued kernel ridge regression from the input kernel $K$ to the SDR representation, providing a stable and well-posed projection operator for unseen points. Without loss of generality, we can set $\mu = 1$ since it can be absorbed into the regularization parameter $\lambda_{L}$. In the case that the optimal coupling matrix is a permutation matrix under uniform marginals, the expression for $L^{*}$ simplifies to the usual kernel ridge form,
\begin{equation}
    L^{*} = (K + \lambda_{L}I)^{-1}Z
\end{equation}

\section{FGW Gradient Inertness}
\subsection{Mathematical intuition for the gradient inertness in FGW}
\label{appx:fgw_bottleneck_theory}
In this section, we provide some mathematical intuition into the observation made in Remark \ref{rem:fgw_bottleneck}. \\
Recall the partially minimized BS-FGW objective,
\begin{equation}
    \mathcal{J}(T,Z) = (1 - \alpha)S(T) + \alpha GW(T;Z)
\end{equation}
where $S(T)$ is the supervision loss term defined as,
\begin{equation}
    S(T) = \sum_{i=1}^{n}\sum_{j=1}^{m}L_{s}(y_{i}, g_{j}^{*}(T))T_{ij}
\end{equation}

For part (i), observe that once the prototype targets $g_{j}^{*}(T)$ are replaced with the closed forms from Proposition \ref{prop:proto_targets}, the supervised term $S(T)$ depends only on the coupling matrix $T$ and not explicitly on $Z$. Therefore, for fixed $T$,
\begin{equation}
    \nabla_{Z}S(T) = 0
\end{equation}
The $Z$-update is governed only by the geometric term $\alpha GW(Z;T)$. This establishes that supervision does not directly act on the representation updates.

For Part (ii), let $T^*(Z)$ denote the optimal coupling as a function of $Z$. The joint minimization of $J(Z, T)$ over both variables is equivalent to minimizing the value function
\begin{equation}
    F(Z) = \underset{T}{\text{min}} \ J(Z, T)
\end{equation}
over Z alone. To compute the gradient of $F(Z)$ through the minimizer, we can invoke the Danskin theorem to show,
\begin{equation}
    \nabla F(Z) = \nabla_{Z}J(Z, T^*(Z))
\end{equation}
wherever the minimiser $T^*$ is unique and F is differentiable. This implies, the gradient of the joint objective is equal to the partial gradient with $T$ frozen at its optimal value. From Part (i) we already know the supervised term is purely a function of $T$. Combining this with the fact that $\nabla F(Z)$ does not have a $\nabla_{T}J$ component because $T^*$ is already optimal, we have $\nabla_{Z}S(T) = 0$. Consequently, we can conclude that the supervision term therefore does not contribute to $\nabla F(Z)$ wherever $F$ is differentiable.

For quadratic OT problems like GW, the solution $T^*$ can sit at extreme points of the coupling polytope (\citep{vayer2026extremegw}) and hence, $T^*(Z)$ is piecewise constant in $Z$. This implies $F$ is piecewise-smooth and the supervision gradient is inert on each piece. Therefore, the labels influence $Z$ solely through discrete switches (or flips) of the optimal coupling. This is a non-differentiable gradient channel that cannot be detected by any first-order method like BCD or Gradient Descent. Hence, we conclude that gradient saturation exists almost everywhere except for the points where the optimal coupling on the coupling polytope flips.

\subsection{Empirical Evidence of the Gradient Inertness and the Effect of Explicit Dependence Maximization in SDR}
\label{appx:fgw_bottleneck_evidence}
 In this study, we provide empirical evidence of the gradient inertness and why an explicit dependence maximization term is required for Supervised Distributional Reduction to work. The FGW-only version is a variant which uses the BSFGW objective for optimizing both $\textbf{T}$ and $\textbf{Z}$ with the BCD algorithm. We compare this approach with the SDR objective that includes an explicit dependence maximization term in the form of CKA. We also include DistR and a variant of DistR where we combine the DistR objective with the CKA term to assess what the FGW formulation provides. We report the Homogeneity scores and the Silhouette scores for the DistR, DistR + CKA, FGW-only and SDR variants in Table \ref{tab:fgw_vs_sdr} for the COIL-20 and SNAREseq datasets. 

\begin{table}[h!]
\captionsetup{skip=12pt}
\centering
\resizebox{\textwidth}{!}{
\begin{tabular}{lcc cc cc cc}
\toprule
 & \multicolumn{2}{c}{COIL-20 $(m = 100)$} & \multicolumn{2}{c}{SNAREseq $(m = 100)$}\\
\cmidrule(lr){2-3} \cmidrule(lr){4-5} \cmidrule(lr){6-7}
 & Homogeneity ($\uparrow$) & Silhouette Score ($\uparrow$)  
 & Homogeneity ($\uparrow$) & Silhouette Score ($\uparrow$)   \\
\midrule
DistR & 90.33 $\pm$ 0.50 & 22.96 $\pm$ 9.35 & 81.62 $\pm$ 0.63 & 30.15 $\pm$ 3.24 \\
DistR + CKA & 90.35 $\pm$ 0.51 & 51.63 $\pm$ 1.54 & 81.79 $\pm$ 0.36 & 74.22 $\pm$ 9.67 \\
FGW-only & 99.71 $\pm$ 0.32 & 32.04 $\pm$ 2.26 & 96.01 $\pm$ 1.52 & 13.43 $\pm$ 1.63 \\
\textbf{SDR} & 99.27 $\pm$ 0.61 & 55.65 $\pm$ 2.15 & 92.46 $\pm$ 0.70 & 80.05 $\pm$ 8.17 \\
\bottomrule
\end{tabular}}
\caption{Homogeneity and Silhouette Score comparison for different objectives across 5 seeds}
\label{tab:fgw_vs_sdr}
\end{table}

\begin{figure}[htbp]
\centering

    \begin{subfigure}{0.45\textwidth}
        \includegraphics[width=\linewidth]{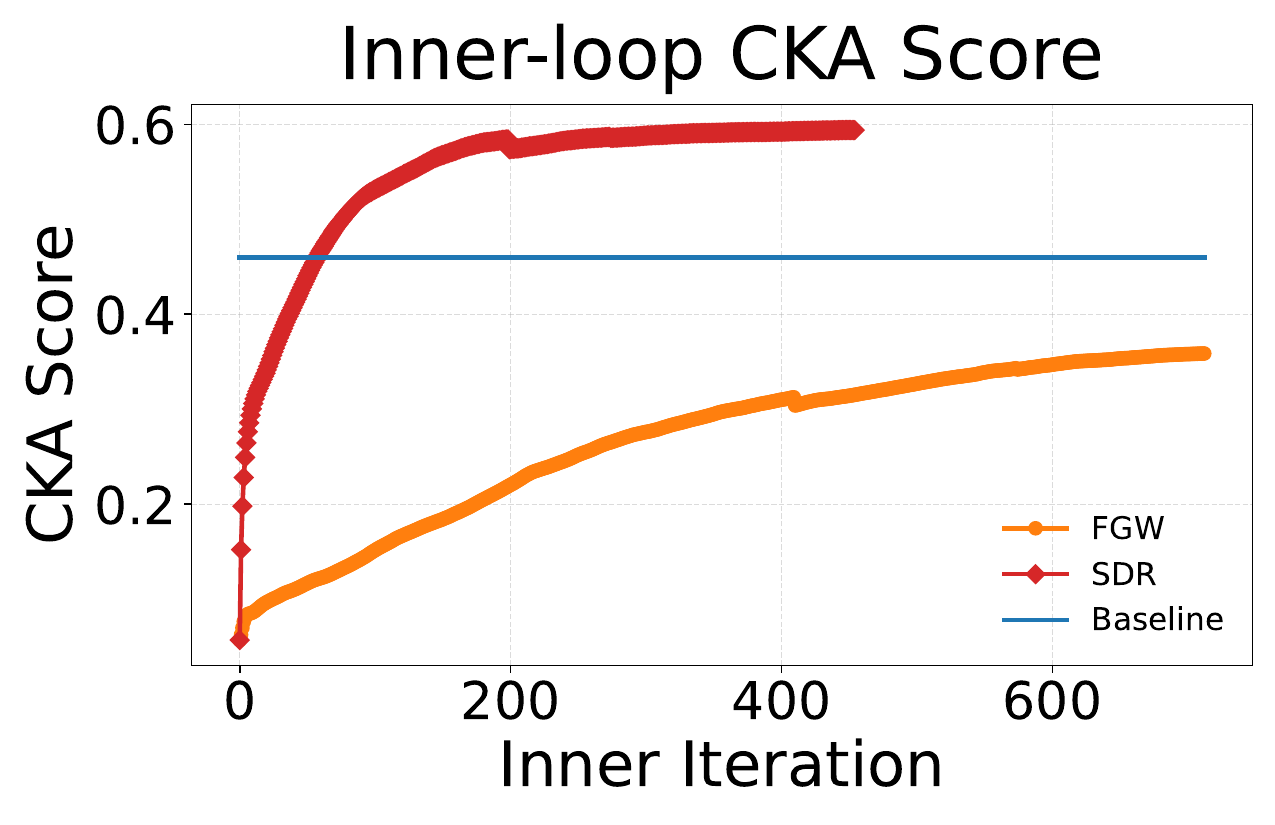}
        \caption{Typical first inner loop for COIL-20}
    \end{subfigure}
    \hfill
    \begin{subfigure}{0.45\textwidth}
        \includegraphics[width=\linewidth]{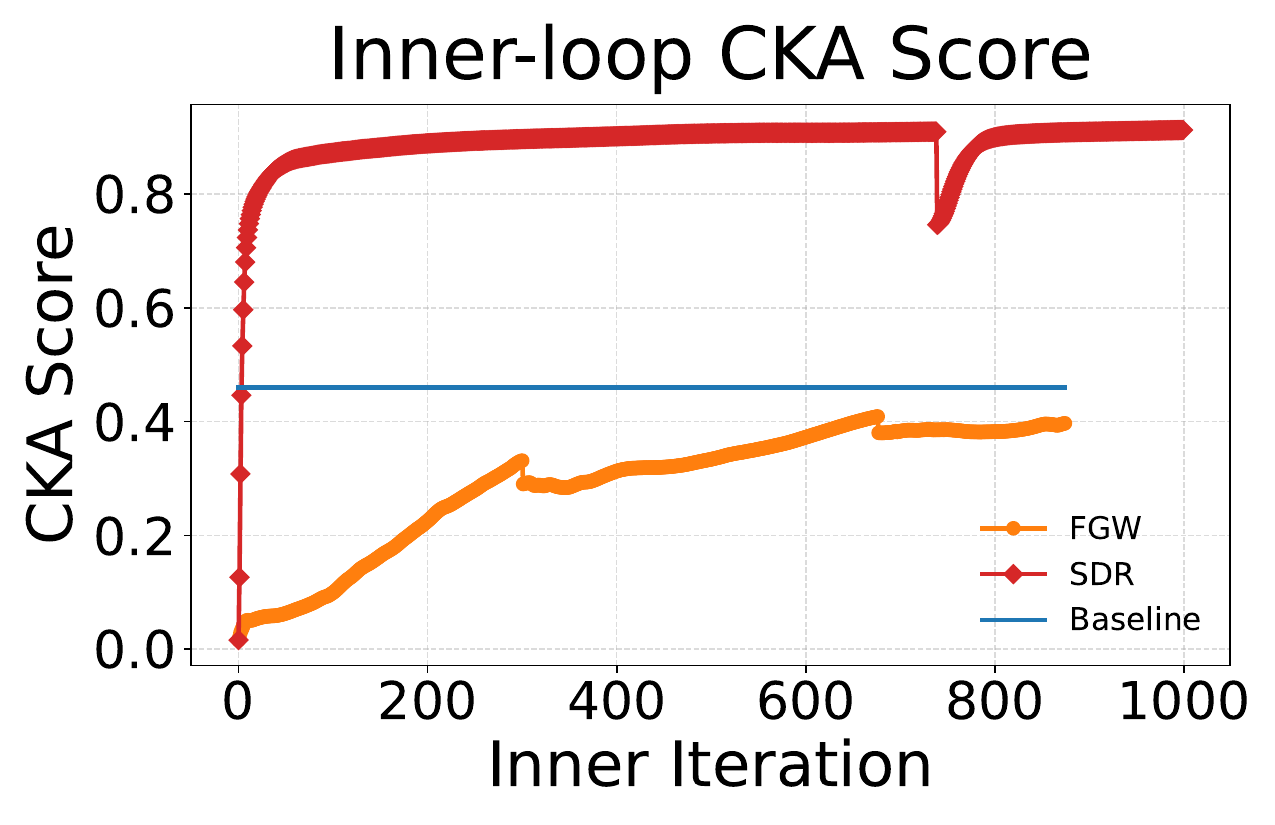}
        \caption{Typical first inner loop for SNARESeq}
    \end{subfigure}
    
    \vspace{0.5cm}
    
    \begin{subfigure}{0.45\textwidth}
        \includegraphics[width=\linewidth]{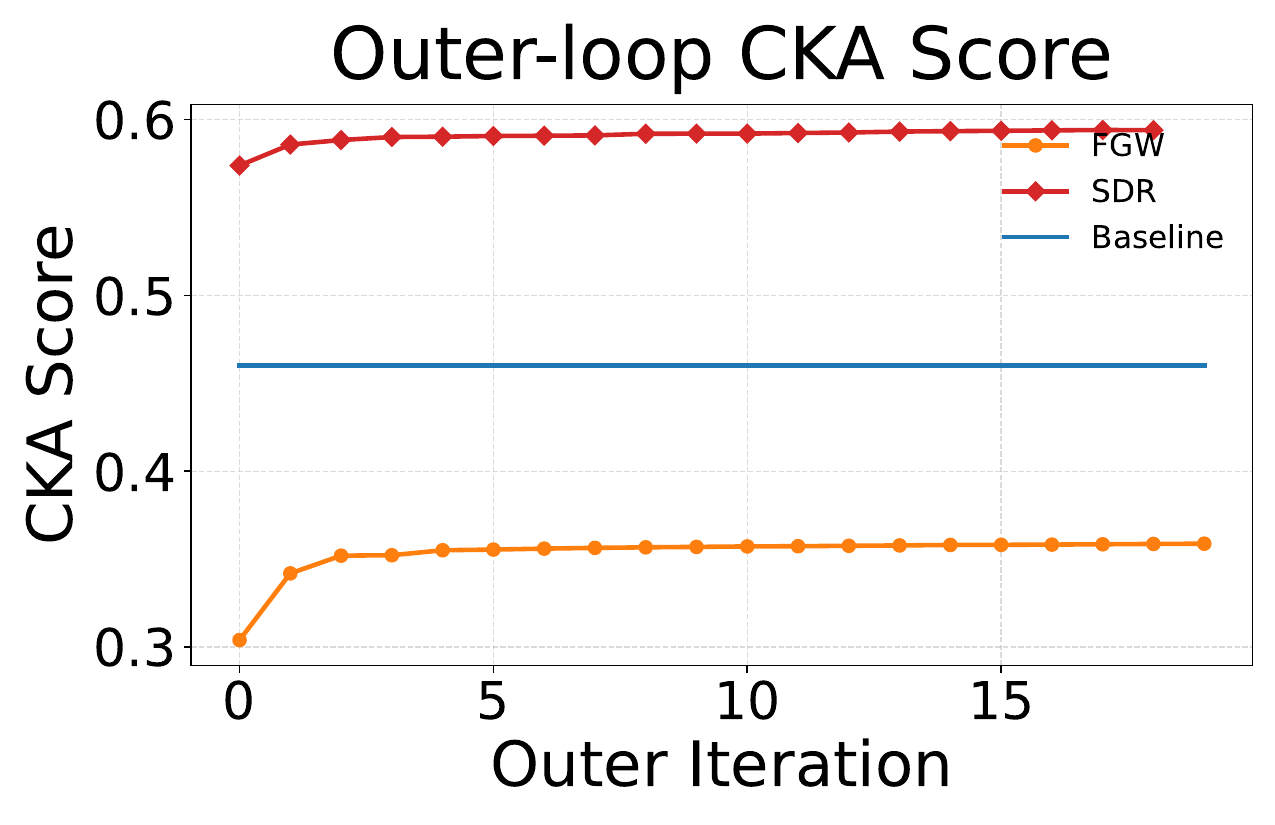}
        \caption{Typical Outer Loops for COIL-20}
    \end{subfigure}
    \hfill
    \begin{subfigure}{0.45\textwidth}
        \includegraphics[width=\linewidth]{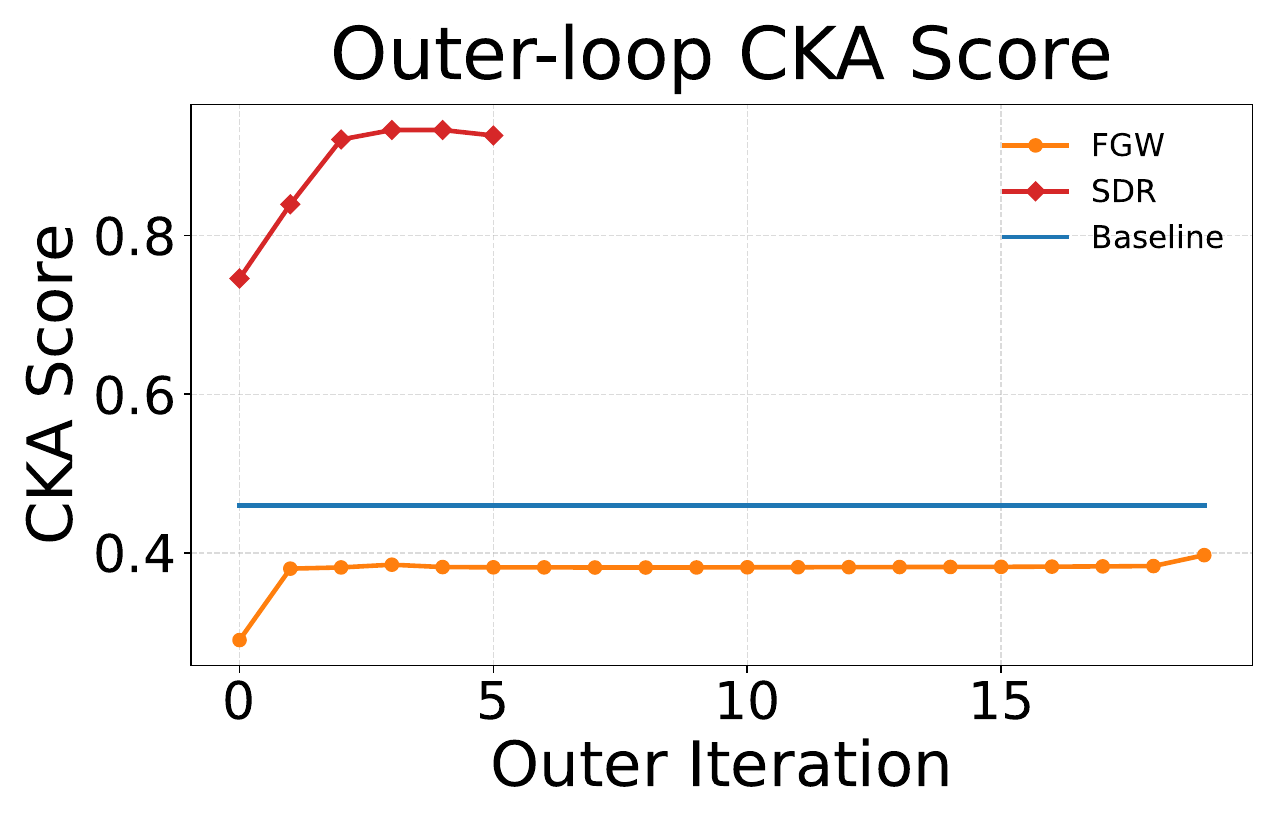}
        \caption{Typical Outer Loops for SNARESeq}
    \end{subfigure}

\caption{Variation of the CKA score across inner and outer loops of the optimization process for COIL20 and SNARESeq datasets.}
\label{fig:cka_scores}
\end{figure}

\begin{figure}[h!]
    \centering
    \includegraphics[width=0.83\linewidth]{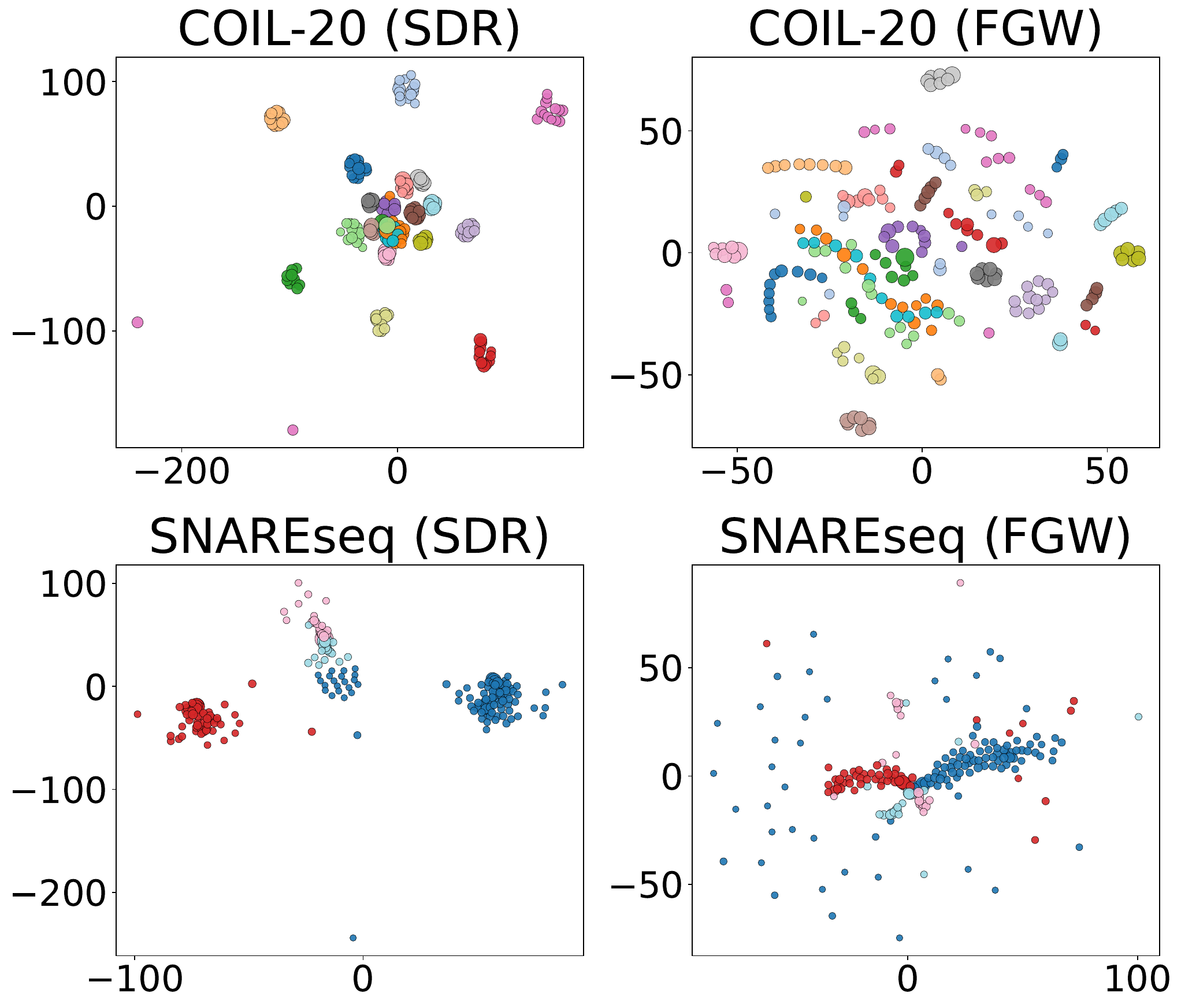}
    \caption{Example embeddings for the COIL20 and SNAREseq datasets for the SDR and FGW-only objectives for visual comparison.}
    \label{fig:viz_embeddings}
\end{figure}

 We can observe from Table \ref{tab:fgw_vs_sdr} that the DistR + CKA variant improves on the Silhouette score but not on the Homogeneity score as the optimization of $T$ is unsupervised and this does not improve cluster purity. On the other hand, both FGW-only and the SDR formulations that use FGW for the $T$-step improve on the Homogeneity scores. On the other axis of the analysis, the FGW-only formulation does not improve over DistR when it comes to the silhouette score because of the gradient inertness discussed in Remark \ref{rem:fgw_bottleneck}. Since, SDR incorporates target information through both channels - the FGW supervision term and through the CKA term, it improves on both metrics compared to its DistR ad FGW counterparts.

Plots (a) and (b) in Figure \ref{fig:cka_scores} show how the CKA score of the embeddings $Z$ with respect to the targets evolve in the first inner loop of the SDR objective. This inner loop corresponds to the optimization of $Z$ using an Adam optimizer and runs at most for 1000 steps. Plots (c) and (d) show how the CKA score progresses with the outer loop which consists of two steps: an inner loop for optimizing $Z$ and a solve of the srBSFGW objective to update the coupling matrix $T$. Clearly, the FGW objective plateaus and produces a worse CKA score than the actual data, whereas the SDR objective produces a high CKA score indicating the retention of predictive signal leading to target-aware embeddings.

Figure \ref{fig:viz_embeddings} shows example embeddings generated by SDR and the FGW-only variant for the COIL-20 and SNAREseq datasets. SDR produces reasonably well separated clusters of the different classes in the dataset whereas the FGW-only variant while managing to bring prototypes from the same class together, does not form well separated clusters. These results indicate that the FGW extension is useful but the dependence maximization term in the SDR objective makes it much more effective in capturing the nuances of the dataset.

\section{Distributional Reduction Experimental Details}
\label{appx:distr_expts}
\subsection{Datasets}
The experiments for distributional reduction (joint clustering and DR) used three datasets with varying properties. We use the COIL-20 \citep{Nene1996coil20}, Fashion MNIST \citep{xiao2017fmnist} and SNAREseq \citep{chen2019snareseq} datasets for all distributional reduction experiments. The details of these datasets are provided in Table \ref{tab:distr_datasets}.

\begin{table}[h!]
\captionsetup{skip=10pt}
    \centering
    \begin{tabular}{|c|c|c|c|}
        \hline
         \textbf{Dataset} & \textbf{Number of Points} & \textbf{Input Features} & \textbf{Number of Classes} \\
         \hline
         COIL-20 & 1440 & 16384 & 20 \\
         \hline
         Fashion-MNIST & 10000 & 784 & 10 \\
         \hline
         SNAREseq (Chromatin) & 1047 & 19 & 4 \\
         \hline
    \end{tabular}
    \caption{Details of Datasets used in Distributional Reduction experiments}
    \label{tab:distr_datasets}
\end{table}

COIL-20 and the Fashion-MNIST datasets were pre-processed with PCA to reduce the number of features to 50 as done in practice (\cite{vanassel2025distr}, \cite{vandermaaten2008tsne}) and the inputs were normalized for all three datasets.

\subsection{Experiments}
The distributional reduction experiments were performed on 3 different datasets as mentioned in the previous sub-section. The methods tested were DistR, Cluster-then-DR and DR-then-Cluster following the same experimental setup in \cite{vanassel2025distr}. The experiments were run for different number of prototypes \{20, 60, 100, 140, 180\} with 5 random seeds. For evaluation metrics, we consider the Homogeneity score, k-means Normalized Mutual Information and the Silhouette Score as defined in \cite{vanassel2025distr}. All experiments were run on a remote server equipped with an NVIDIA-L4 GPU.

\subsection{Hyperparameter settings}
All steps involving clustering both for the initialization of the coupling matrices and for methods that explicitly perform clustering, the Spectral Clustering implementation from scikit-learn was used. For all experiments, DistR and SDR used the Symmetric Entropic Affinity \citep{vanassel2023sea} for $\textbf{X}$ with the perplexity hyperparameter tuned between \{20, 30, 50, 100, 200\} and the learning rate tuned between \{0.01, 0.1, 1.0\}. The learning rates were chosen as the highest value in the set that did not break the optimization routine of the SEA transport problem. We report these tuned values for the various datasets in Table \ref{tab:sea_params}.

\begin{table}[h!]
\captionsetup{skip=10pt}
    \centering
    \begin{tabular}{|c|c|c|}
        \hline
         \textbf{Dataset} & \textbf{Perplexity} & \textbf{Learning Rate} \\
         \hline
         COIL-20 & 20 & 0.1 \\
         \hline
         Fashion-MNIST & 50 & 0.01 \\
         \hline
         SNAREseq (Chromatin) & 30 & 0.01 \\
         \hline
    \end{tabular}
    \caption{Hyperparameter details of SEA for various datasets}
    \label{tab:sea_params}
\end{table}

For the embeddings $\textbf{Z}$, the normalized Student-t affinity \citep{vandermaaten2008tsne} was used. For the CKA computation, we require positive semi-definite kernels and so affinities cannot be directly used. Therefore, we choose the kernel that is closest in behaviour to the normalized Student-t affinity which is the Rational Quadratic (RQ) kernel \citep{Rasmussen2006Gaussian} given by
\begin{equation}
    k_{RQ}(x, x') = \left(1 + \frac{||x - x'||^{2}}{2 \alpha l^{2}}\right)^{-\alpha}
\end{equation}

This kernel is equivalent to adding several SE kernels with different lengthscale parameters. The parameter $\alpha$ determines how large scale variations are weighted relative to small scale variations. The RQ kernel behaves like a heavy-tailed similarity function since it decays polynomially. As all the datasets used in these experiments are classification datasets, we use the delta kernel for the outputs as described in \cite{barshan2011spca}. For all experiments, $\alpha$ in the srBSFGW objective was set to 0.2 and $\eta$ in the SDR objective was set to 1000. The Adam optimizer was used for updating the embeddings $Z$ with the default learning rate of 0.001.

\section{Accept-Reject step for SDR}
\label{appx:safeguard}
In this section, we show some results on the safeguarded variant of SDR. This modification of SDR introduces an accept-reject step: a $T$-update is accepted only if it does not increase the full objective $J$, otherwise the previous coupling matrix is retained. It should be noted that, accepted steps are non-increasing and guarantee monotonic descent on the $T$-block. We compare this variant with the original algorithm in Table \ref{tab:sdr_vs_sdrar} where the experiments were performed on the COIL-20 and SNAREseq datasets for 3 random seeds and the results averaged across $m \in 20, 60, 100, 140, 180$.

\begin{table}[h!]
\captionsetup{skip=12pt}
\centering
\resizebox{\textwidth}{!}{
\begin{tabular}{lccc ccc}
\toprule
 Method & \multicolumn{3}{c}{COIL-20} & \multicolumn{3}{c}{SNAREseq}\\
\cmidrule(lr){2-4} \cmidrule(lr){5-7}
 & Homogeneity ($\uparrow$) & k-Means NMI ($\uparrow$) & Silhouette Score ($\uparrow$)  
 & Homogeneity ($\uparrow$) & k-Means NMI ($\uparrow$) & Silhouette Score ($\uparrow$)   \\
\midrule
SDR & 94.95 $\pm$ 7.93 & 91.69 $\pm$ 3.01 & 48.00 $\pm$ 17.01 & 93.35 $\pm$ 1.53 & 90.69 $\pm$ 2.53 & 82.66 $\pm$ 7.28 \\
SDR gated on $J$ & 94.95 $\pm$ 7.93 & 91.69 $\pm$ 3.01 & 48.00 $\pm$ 17.01 & 90.62 $\pm$ 2.22 & 87.16 $\pm$ 4.77 & 80.36 $\pm$ 8.41 \\
\bottomrule
\end{tabular}}
\caption{Comparison of vanilla SDR and SDR with an accept-reject gate on $J$.}
\label{tab:sdr_vs_sdrar}
\end{table}
The safeguarded variant is indistinguishable from our original scheme on COIL-20 — the FGW $T$-update acts as a descent step for $J$ in practice. On SNAREseq it performs slightly worse, with the gap growing with $m$ as shown in Table \ref{tab:safeguard_m}.

\begin{table}[h!]
\captionsetup{skip=12pt}
\centering
\resizebox{\textwidth}{!}{
\begin{tabular}{lccc ccc}
\toprule
 m & \multicolumn{3}{c}{SDR} & \multicolumn{3}{c}{SDR gated on J}\\
\cmidrule(lr){2-4} \cmidrule(lr){5-7}
 & Homogeneity ($\uparrow$) & k-Means NMI ($\uparrow$) & Silhouette Score ($\uparrow$)  
 & Homogeneity ($\uparrow$) & k-Means NMI ($\uparrow$) & Silhouette Score ($\uparrow$)   \\
\midrule
20 & 93.84 $\pm$ 3.90 & 94.05 $\pm$ 3.70 & 93.19 $\pm$ 3.85 & 93.84 $\pm$ 3.90 & 94.05 $\pm$ 3.70 & 93.19 $\pm$ 3.85 \\ 
60 & 91.27 $\pm$ 0.44 & 91.44 $\pm$ 0.32 & 85.52 $\pm$ 1.51 & 90.48 $\pm$ 0.79 & 90.07 $\pm$ 1.06 & 83.95 $\pm$ 1.35 \\
100 & 92.90 $\pm$ 0.05 & 89.09 $\pm$ 5.12 & 79.79 $\pm$ 11.54 & 91.43 $\pm$ 0.63 & 84.54 $\pm$ 5.02 & 75.31 $\pm$ 13.70 \\
140 & 93.23 $\pm$ 0.13 & 91.44 $\pm$ 1.36 & 81.19 $\pm$ 1.51 & 87.95 $\pm$ 7.40 & 84.76 $\pm$ 10.18 & 77.38 $\pm$ 4.95 \\
180 & 95.50 $\pm$ 1.66 & 87.43 $\pm$ 8.51 & 73.58 $\pm$ 13.79 & 89.37 $\pm$ 8.05 & 82.40 $\pm$ 11.20 & 71.94 $\pm$ 12.64 \\
\bottomrule
\end{tabular}}
\caption{Comparison of vanilla SDR with the safeguarded variant gated on the objective $J$.}
\label{tab:safeguard_m}
\end{table}

The accept-reject mechanism has only one action - to reject when the objective increases from a $T$-update. This can force the optimization routine to be stuck in a loop. The occasional non-monotone $T$-steps whose magnitude is $\eta$-bounded, controls and helps the scheme escape poor local minima, which is very common in non-convex alternating minimization. An alternative to this would be to use line search/backtracking methods instead of having a simple accept-reject gate. We leave this improvements and convergence guarantees as future work.

\section{Comparison with Sequential Baselines}
The core SDR method brings two data reduction paradigms together namely: Clustering and Dimensionality Reduction (DR). SDR produces prototypes by jointly performing clustering and DR. In these ablations, we test how the joint reduction procedure compares with sequential baselines, where clustering and DR are performed in a sequence, one after another. We tested two protocols:
\begin{itemize}
    \item \textbf{Protocol A} - Clustering and DR both steps are supervised. Both stages run SDR, clustering step runs with $m < n$ and the DR step runs with $m = n$.
    \item \textbf{Protocol B} - Clustering step is unsupervised and uses spectral clustering, but the DR step is supervised and runs SDR with $m = n$.
\end{itemize}
Protocol A is, in essence closer to the SDR algorithm as both stages are supervised, whereas Protocol B is closer to the DistR evaluation framework. Since both variants use SDR as part of the clustering and DR steps, we use this ablation to understand which mode of usage of SDR works better for the datasets we used in the experiments. The results for $m = 20, 100, 200$ over 5 random seeds are provided in Tables \ref{tab:sb_m20}, \ref{tab:sb_m100} and \ref{tab:sb_m200} respectively for the COIL-20 and SNAREseq datasets.

\begin{table}[h!]
\captionsetup{skip=12pt}
\centering
\resizebox{\textwidth}{!}{
\begin{tabular}{lcccc cccc}
\toprule
 Method & \multicolumn{4}{c}{COIL-20 (m=20)} & \multicolumn{4}{c}{SNAREseq (m = 20)}\\
\cmidrule(lr){2-5} \cmidrule(lr){6-9}
 & Homogeneity ($\uparrow$) & k-Means NMI ($\uparrow$) & Silhouette Score ($\uparrow$) & Runtime (s) ($\downarrow$)
 & Homogeneity ($\uparrow$) & k-Means NMI ($\uparrow$) & Silhouette Score ($\uparrow$) & Runtime (s) ($\downarrow$)  \\
\midrule
Spec-C$\rightarrow$DR & 60.33 $\pm$ 0.09 & 68.55 $\pm$ 0.06 & 21.70 $\pm$ 2.97 & 3.03 $\pm$ 0.15 & 76.45 $\pm$ 0.00 & 66.46 $\pm$ 10.09 & 53.94 $\pm$ 7.84 & 1.77 $\pm$ 0.15 \\
DR$\rightarrow$Spec-C & 89.13 $\pm$ 4.89 & 91.57 $\pm$ 3.24 & 6.03 $\pm$ 2.93 & 209.24 $\pm$ 87.32 & 81.38 $\pm$ 2.02 & 77.91 $\pm$ 4.74 & 65.69 $\pm$ 2.39 & 113.07 $\pm$ 26.64 \\
C$\rightarrow$DR & 81.07 $\pm$ 0.00 & 86.52 $\pm$ 0.00 & 24.26 $\pm$ 0.44 & 3.39 $\pm$ 0.15 & 100.00 $\pm$ 0.00 & 95.99 $\pm$ 8.96 & 87.95 $\pm$ 12.73 & 1.98 $\pm$ 0.06 \\
DR$\rightarrow$C & 88.85 $\pm$ 6.18 & 91.81 $\pm$ 3.96 & 5.65 $\pm$ 5.10 & 214.51 $\pm$ 86.21 & 73.79 $\pm$ 19.27 & 72.54 $\pm$ 20.51 & 60.18 $\pm$ 23.76 & 115.88 $\pm$ 26.20 \\
Joint SDR & 81.07 $\pm$ 0.00 & 86.52 $\pm$ 0.00 & 19.42 $\pm$ 3.33 & 3.47 $\pm$ 0.11 & 95.16 $\pm$ 3.40 & 95.31 $\pm$ 3.23 & 92.24 $\pm$ 4.50 & 2.49 $\pm$ 0.83 \\
\bottomrule
\end{tabular}}
\caption{Comparison with sequential baselines for $m = 20$.}
\label{tab:sb_m20}
\end{table}

\begin{table}[h!]
\captionsetup{skip=12pt}
\centering
\resizebox{\textwidth}{!}{
\begin{tabular}{lcccc cccc}
\toprule
 Method & \multicolumn{4}{c}{COIL-20 (m=100)} & \multicolumn{4}{c}{SNAREseq (m = 100)}\\
\cmidrule(lr){2-5} \cmidrule(lr){6-9}
 & Homogeneity ($\uparrow$) & k-Means NMI ($\uparrow$) & Silhouette Score ($\uparrow$) & Runtime (s) ($\downarrow$)
 & Homogeneity ($\uparrow$) & k-Means NMI ($\uparrow$) & Silhouette Score ($\uparrow$) & Runtime (s) ($\downarrow$)  \\
\midrule
Spec-C$\rightarrow$DR & 90.16 $\pm$ 0.63 & 78.52 $\pm$ 2.62 & 37.68 $\pm$ 6.99 & 3.31 $\pm$ 0.16 & 79.32 $\pm$ 1.95 & 68.93 $\pm$ 4.74 & 56.63 $\pm$ 7.15 & 2.25 $\pm$ 0.38 \\
DR$\rightarrow$Spec-C & 87.72 $\pm$ 4.30 & 84.93 $\pm$ 3.72 & 51.44 $\pm$ 7.19 & 206.27 $\pm$ 85.57 & 70.50 $\pm$ 4.25 & 67.22 $\pm$ 5.56 & 52.00 $\pm$ 10.88 & 112.84 $\pm$ 26.23 \\
C$\rightarrow$DR & 98.78 $\pm$ 0.31 & 93.02 $\pm$ 2.40 & 68.79 $\pm$ 3.56 & 4.39 $\pm$ 0.83 & 93.06 $\pm$ 0.84 & 93.06 $\pm$ 0.76 & 89.45 $\pm$ 1.55 & 3.69 $\pm$ 0.66 \\
DR$\rightarrow$C & 92.26 $\pm$ 7.68 & 90.20 $\pm$ 5.73 & 32.62 $\pm$ 37.72 & 213.44 $\pm$ 86.62 & 79.88 $\pm$ 5.79 & 30.50 $\pm$ 42.39 & 34.37 $\pm$ 35.39 & 118.12 $\pm$ 22.71 \\
Joint SDR & 99.27 $\pm$ 0.61 & 94.10 $\pm$ 1.54 & 55.65 $\pm$ 2.15 & 4.36 $\pm$ 0.28 & 92.46 $\pm$ 0.70 & 89.42 $\pm$ 3.69 & 80.05 $\pm$ 8.17 & 3.44 $\pm$ 0.39 \\
\bottomrule
\end{tabular}}
\caption{Comparison with sequential baselines for $m = 100$.}
\label{tab:sb_m100}
\end{table}

\begin{table}[h!]
\captionsetup{skip=12pt}
\centering
\resizebox{\textwidth}{!}{
\begin{tabular}{lcccc cccc}
\toprule
 Method & \multicolumn{4}{c}{COIL-20 (m=200)} & \multicolumn{4}{c}{SNAREseq (m = 200)}\\
\cmidrule(lr){2-5} \cmidrule(lr){6-9}
 & Homogeneity ($\uparrow$) & k-Means NMI ($\uparrow$) & Silhouette Score ($\uparrow$) & Runtime (s) ($\downarrow$)
 & Homogeneity ($\uparrow$) & k-Means NMI ($\uparrow$) & Silhouette Score ($\uparrow$) & Runtime (s) ($\downarrow$)  \\
\midrule
Spec-C$\rightarrow$DR & 95.70 $\pm$ 0.42 & 83.88 $\pm$ 2.42 & 50.97 $\pm$ 4.65 & 5.04 $\pm$ 0.81 & 76.59 $\pm$ 1.87 & 71.80 $\pm$ 3.07 & 67.84 $\pm$ 4.90 & 3.64 $\pm$ 0.85 \\
DR$\rightarrow$Spec-C & 84.40 $\pm$ 3.02 & 81.87 $\pm$ 5.01 & 45.47 $\pm$ 4.77 & 211.10 $\pm$ 85.72 & 65.99 $\pm$ 3.01 & 64.80 $\pm$ 2.71 & 40.88 $\pm$ 5.40 & 113.38 $\pm$ 26.64 \\
C$\rightarrow$DR & 96.82 $\pm$ 0.46 & 93.94 $\pm$ 1.59 & 81.24 $\pm$ 3.46 & 7.33 $\pm$ 1.54 & 94.60 $\pm$ 0.80 & 94.67 $\pm$ 0.78 & 90.36 $\pm$ 2.33 & 8.69 $\pm$ 1.04 \\
DR$\rightarrow$C & 62.07 $\pm$ 50.62 & 57.17 $\pm$ 52.24 & 37.50 $\pm$ 39.59 & 217.32 $\pm$ 85.55 & 63.72 $\pm$ 20.64 & 32.89 $\pm$ 45.16 & 24.67 $\pm$ 46.56 & 122.66 $\pm$ 24.35 \\
Joint SDR & 98.86 $\pm$ 0.74 & 91.30 $\pm$ 1.78 & 58.93 $\pm$ 2.70 & 6.61 $\pm$ 0.49 & 94.71 $\pm$ 1.16 & 79.35 $\pm$ 3.45 & 64.13 $\pm$ 9.72 & 7.40 $\pm$ 0.70 \\
\bottomrule
\end{tabular}}
\caption{Comparison with sequential baselines for $m = 200$.}
\label{tab:sb_m200}
\end{table}

Under Protocol A, joint SDR is competitive and is better than most baselines except C$\rightarrow$DR. We hypothesize this is because purity metrics are saturated for any pipeline that clusters first, since supervised clustering in input space can be near-perfect/trivial. DR$\rightarrow$C incurs 30–60× the runtime with prototype collapse and severe seed instability (e.g., silhouette score 34.37 ± 35.39 on SNAREseq for $m=100$).

Under Protocol B, the setting matching the original DistR paper, joint SDR wins Homogeneity and k-Means NMI in every configuration and silhouette in the large majority. The sequential failure modes are complementary: Spec-C$\rightarrow$DR fixes the partition from geometry alone before supervision enters, so target-mixing prototypes can never be repaired downstream; DR$\rightarrow$Spec-C computes the partition without labels on a 2-D projection at 30–60× runtime. Joint optimization is thus the only mechanism transporting label information into both subproblems efficiently.

\section{Hyperparameter Sensitivity - Ablation studies for $\alpha$, $\eta$ and $\beta$}

\subsection{Ablation study ($\beta$): Effect of Projection Regularization}
\label{appx:beta_ablation}
\begin{figure}[h!]
    \centering
    \includegraphics[width=0.82\linewidth]{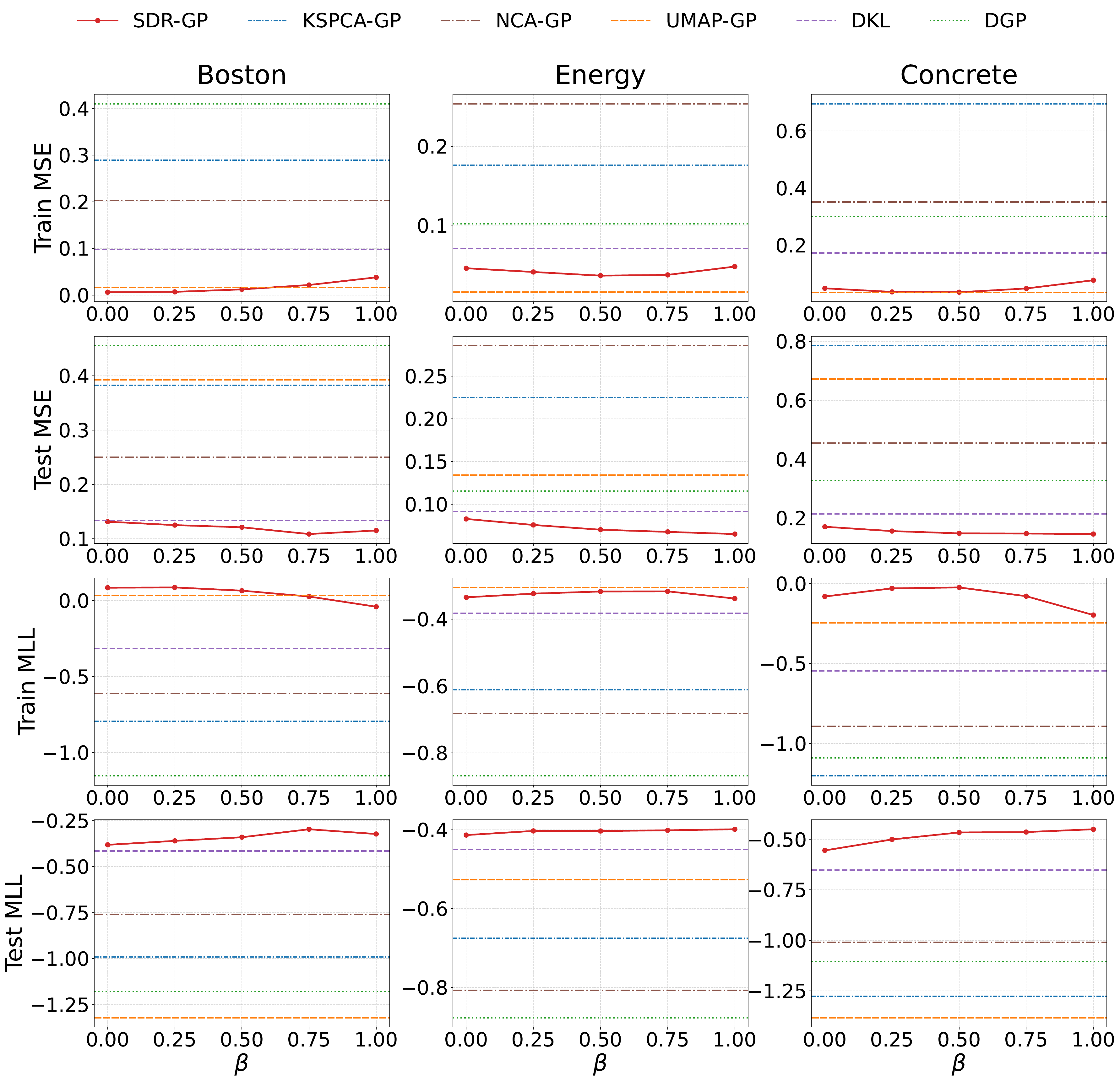}
    \caption{Ablation on the $\beta$ parameter for the regression datasets. A very high $\beta$ value (close to 1) increases training error but improves the test error. A low $\beta$ value (close to 0) decreases the training error but increases the test error. The dashed lines corresponds to the baseline performance of other methods.}
    \label{fig:beta_ablation}
\end{figure}

The hyperparameter $\beta$ used in the SDR-OOS objective for the intermediate projection of the embeddings $Z$ plays a key role in defining the predictive performance of the downstream model. $\beta$ handles the trade-off of the embeddings being representable by a projection map against the retention of predictive signal in the optimized embeddings. In this experiment, we ran a sweep over the range of $\beta$ values where, $\beta \in [0, 1]$. $\beta = 0$ corresponds to no regularization of $Z$ with respect to $L$ and $\beta = 1$ tries to replace the optimized $Z$ in an iteration with its projection onto the RKHS representable set of $K$. We sample $\beta$ at uniform intervals in this range which corresponds to $\{0.0, 0.25, 0.5, 0.75, 1.0\}$ and run the SDR-GP method on all three regression datasets. From the plots it is evident that a high $\beta$ value improves the test error by pulling the optimized $Z$ back onto the RKHS representable set of $K$. However, this can potentially misalign $Z$ with targets $Y$ to some degree and this manifests as increasing training error. On the other hand, a low $\beta$ value improves the training error but increases the test error since the optimization of $Z$ is now less regularized. To maintain a balance between these competing factors, a value close to the mean/median of this range which is 0.5 seems to be a sensible choice for this parameter and consequently we use this value across all other kernel learning experiments. We ran this experiment across 3 different random seeds and the mean is reported in the various plots of Figure \ref{fig:beta_ablation}.

Additionally, the behavior observed across the $\beta$ sweep highlights the role of the SDR regularization term as an implicit complexity control mechanism on the learned embeddings. The relatively stable performance observed around intermediate $\beta$ values further suggests that SDR-GP is not overly sensitive to precise tuning of this parameter, indicating robustness of the proposed objective across a reasonably broad operating regime. This behavior is consistent across all evaluated datasets and supports the interpretation of $\beta$ as governing a bias-variance trade-off between representability and predictive alignment of the optimized embeddings.

\subsection{Ablation study ($\eta$): Effect of Dependence Strength}
\label{appx:eta_ablation}
\begin{figure}[h!]
    \centering
    \includegraphics[width=0.95\linewidth]{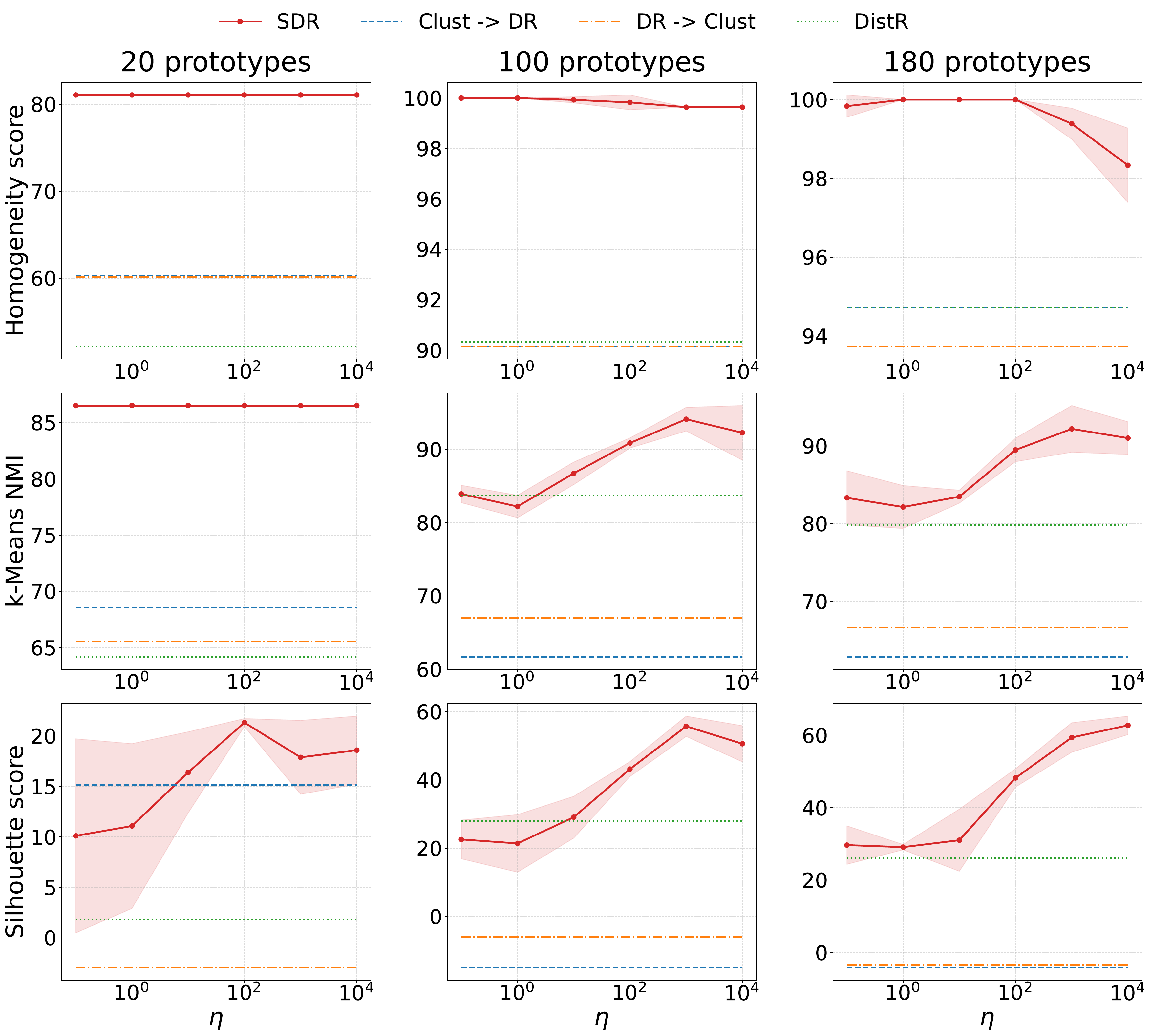}
    \caption{Scores (x100) across varying values of $\eta$ for the COIL-20 dataset averaged over three random seeds. The flat lines show the average performance of the other methods as a baseline for comparison for a given number of prototypes.}
    \label{fig:eta_coil20}
    \end{figure}

\begin{figure}[h!]
    \centering
    \includegraphics[width=1.0\linewidth]{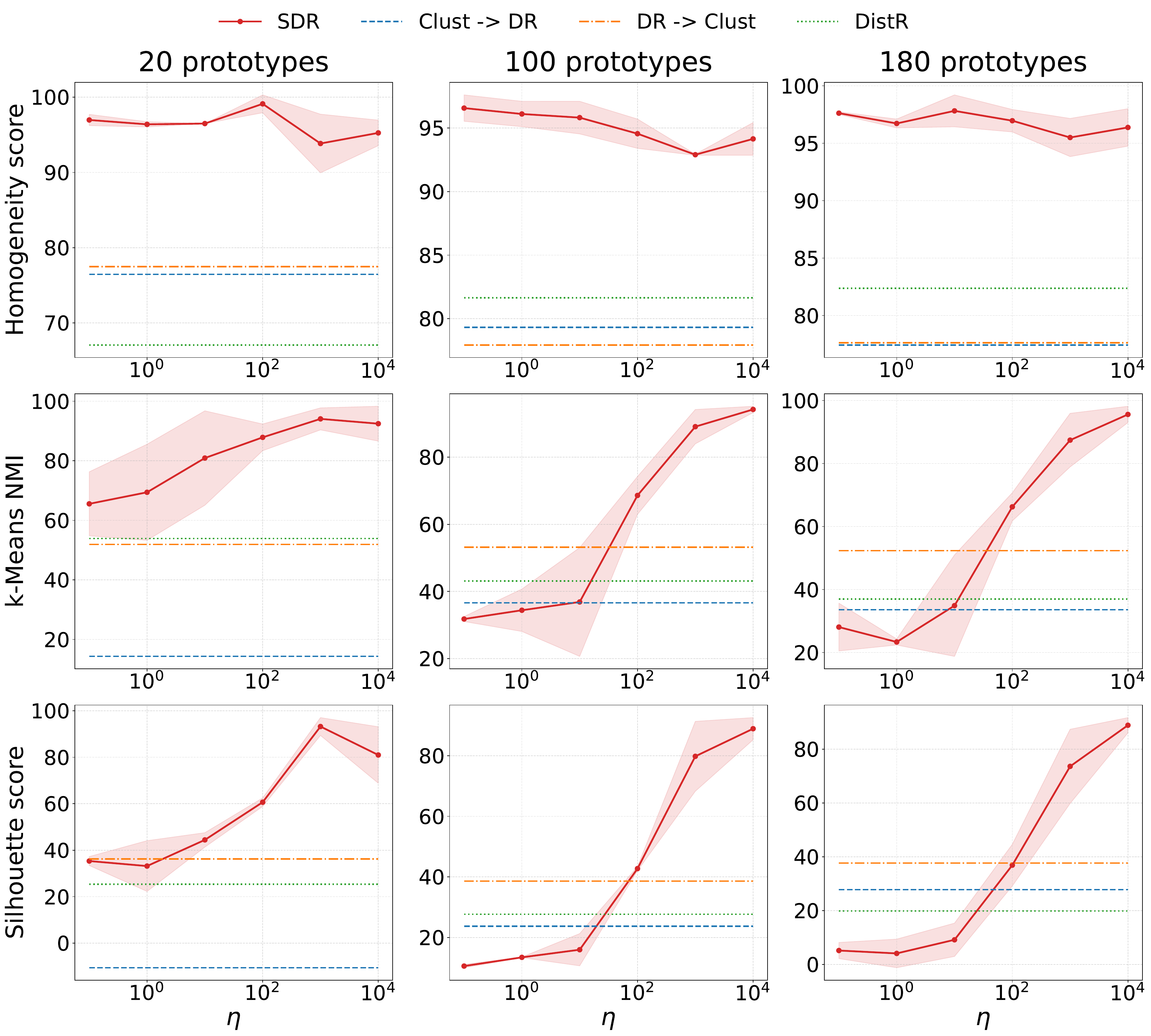}
    \caption{Scores (x100) across varying values of $\eta$ for the SNAREseq dataset averaged over three random seeds. The flat lines show the average performance of the other methods as a baseline for comparison for a given number of prototypes.}
    \label{fig:eta_snareseq}
\end{figure}

$\eta$ is a very important hyperparameter that determines the influence of the targets in the optimization process of the embeddings $Z$ according to the SDR objective. In this section, we present some results on the ablation of the $\eta$ parameter. Since the CKA term is a normalized version of the HSIC loss it always lies in $[0,1]$. Therefore, depending on the magnitude of the GW loss $\eta$ can be suitably tuned. We sweep across $\eta$ in the log space and for three different number of prototypes \{20, 100, 180\}. The results for the COIL-20 and SNAREseq datasets are presented in Figures \ref{fig:eta_coil20} and \ref{fig:eta_snareseq} respectively. For most of the cases in Figure \ref{fig:eta_coil20} and Figure \ref{fig:eta_snareseq}, it is evident that a higher $\eta$ value improves the quality of the prototypes produced by SDR. However, we observe diminishing gains in the quality of the embeddings in many of these plots after a value of $10^{3}$ for $\eta$. Also, for the DKL experiments there was a trade-off that we observed between how well the OOS map $L$ can approximate a very complex map produced by SDR through OT and maximizing predictive signal in the prototypes. Therefore, a reasonably high value of $\eta$ produces high-quality prototypes and $\eta$ was set to $10^3$ for all our experiments.

\subsection{Ablation study ($\alpha$): Effect of Supervision on $T$}
\label{appx:alpha_ablation}
As stated in \cite{vayer2018fgw}, the interpretation of $\alpha$ is to interpolate between prioritising feature similarity (output similarity for SDR) and relational structure. In this work, we consider metrics that look at purity of the formed clusters and retention of predictive fidelity since that is the focus of this work. Under these metrics, the ablation over $\alpha$ shows an expected trend - worst predictive alignment for $\alpha \approx 1$ and best predictive alignment for $\alpha \approx 0$. We ran SDR with $\alpha \in \{0.10, 0.25, 0.50, 0.75, 0.90\}$ for the COIL-20 and SNAREseq datasets with $m = 100$. We report the results for the two datasets in Tables \ref{tab:alpha_coil20} and \ref{tab:alpha_snareseq}.

\begin{table}[h!]
\captionsetup{skip=12pt}
\centering
\resizebox{\textwidth}{!}{
\begin{tabular}{lccccc}
\toprule
$\alpha$ & Homogeneity ($\uparrow$) & k-Means NMI ($\uparrow$) & Silhouette Score ($\uparrow$) & CKA ($\uparrow$) & Runtime (s) ($\downarrow$) \\
\midrule
0.10 & 100.00 $\pm$ 0.00 & 93.17 $\pm$ 0.32 & 55.73 $\pm$ 3.68 & 0.617 $\pm$ 0.008 & 4.98 \\
0.25 & 96.61 $\pm$ 0.06 & 89.81 $\pm$ 4.84 & 49.14 $\pm$ 6.57 & 0.601 $\pm$ 0.013 & 4.05 \\
0.50 & 91.12 $\pm$ 0.83 & 90.98 $\pm$ 1.47 & 50.24 $\pm$ 6.04 & 0.610 $\pm$ 0.008 & 5.11 \\
0.75 & 90.46 $\pm$ 0.57 & 88.20 $\pm$ 1.74 & 43.66 $\pm$ 3.22 & 0.584 $\pm$ 0.007 & 5.10 \\
0.90 & 90.34 $\pm$ 0.49 & 87.55 $\pm$ 2.17 & 46.60 $\pm$ 4.01 & 0.589 $\pm$ 0.026 & 5.58 \\
\bottomrule
\end{tabular}}
\caption{Ablation of the SDR interpolation parameter $\alpha$ on COIL-20. Mean $\pm$ std over seeds 0--2 (perplexity 20, learning rate 0.1).}
\label{tab:alpha_coil20}
\end{table}

\begin{table}[h!]
\captionsetup{skip=12pt}
\centering
\resizebox{\textwidth}{!}{
\begin{tabular}{lccccc}
\toprule
$\alpha$ & Homogeneity ($\uparrow$) & k-Means NMI ($\uparrow$) & Silhouette Score ($\uparrow$) & CKA ($\uparrow$) & Runtime (s) ($\downarrow$) \\
\midrule
0.10 & 98.92 $\pm$ 0.16 & 88.22 $\pm$ 10.19 & 81.96 $\pm$ 13.15 & 0.916 $\pm$ 0.019 & 3.84 \\
0.25 & 89.68 $\pm$ 0.34 & 88.87 $\pm$ 0.54 & 81.57 $\pm$ 6.14 & 0.925 $\pm$ 0.003 & 3.44 \\
0.50 & 86.44 $\pm$ 0.73 & 78.03 $\pm$ 4.23 & 64.49 $\pm$ 4.95 & 0.889 $\pm$ 0.022 & 3.34 \\
0.75 & 84.51 $\pm$ 0.78 & 72.40 $\pm$ 3.50 & 60.65 $\pm$ 1.78 & 0.909 $\pm$ 0.008 & 3.19 \\
0.90 & 82.93 $\pm$ 1.04 & 72.90 $\pm$ 4.05 & 60.06 $\pm$ 2.18 & 0.892 $\pm$ 0.013 & 3.22 \\
\bottomrule
\end{tabular}}
\caption{Ablation of the SDR interpolation parameter $\alpha$ on SNAREseq. Mean $\pm$ std over seeds 0--2 (perplexity 30, learning rate 0.01).}
\label{tab:alpha_snareseq}
\end{table}

Depending on the goal of the reduction task, $\alpha$ can be varied between 0 and 1 to preserve different properties in the embeddings. Since, the goal of this work is retain predictive fidelity, small values of  $\alpha$ are preferred to retain maximum predictive fidelity. It is worth noting that GW is still performing its role of structure preservation, but we do not measure metrics related to geometric preservation here and thus the trend is seemingly monotonous. The discrepancies between $\alpha = 0.25$ and $\alpha=0.5$ for the COIL-20 dataset highlight that the trend is non-monotonous, showing that the GW term is battling with target alignment in that range.

\section{Additional DR Visualizations and Downstream Evaluation}
\label{appx:dr_results}
\begin{figure}[htbp]
    \centering
    \includegraphics[width=1.0\linewidth]{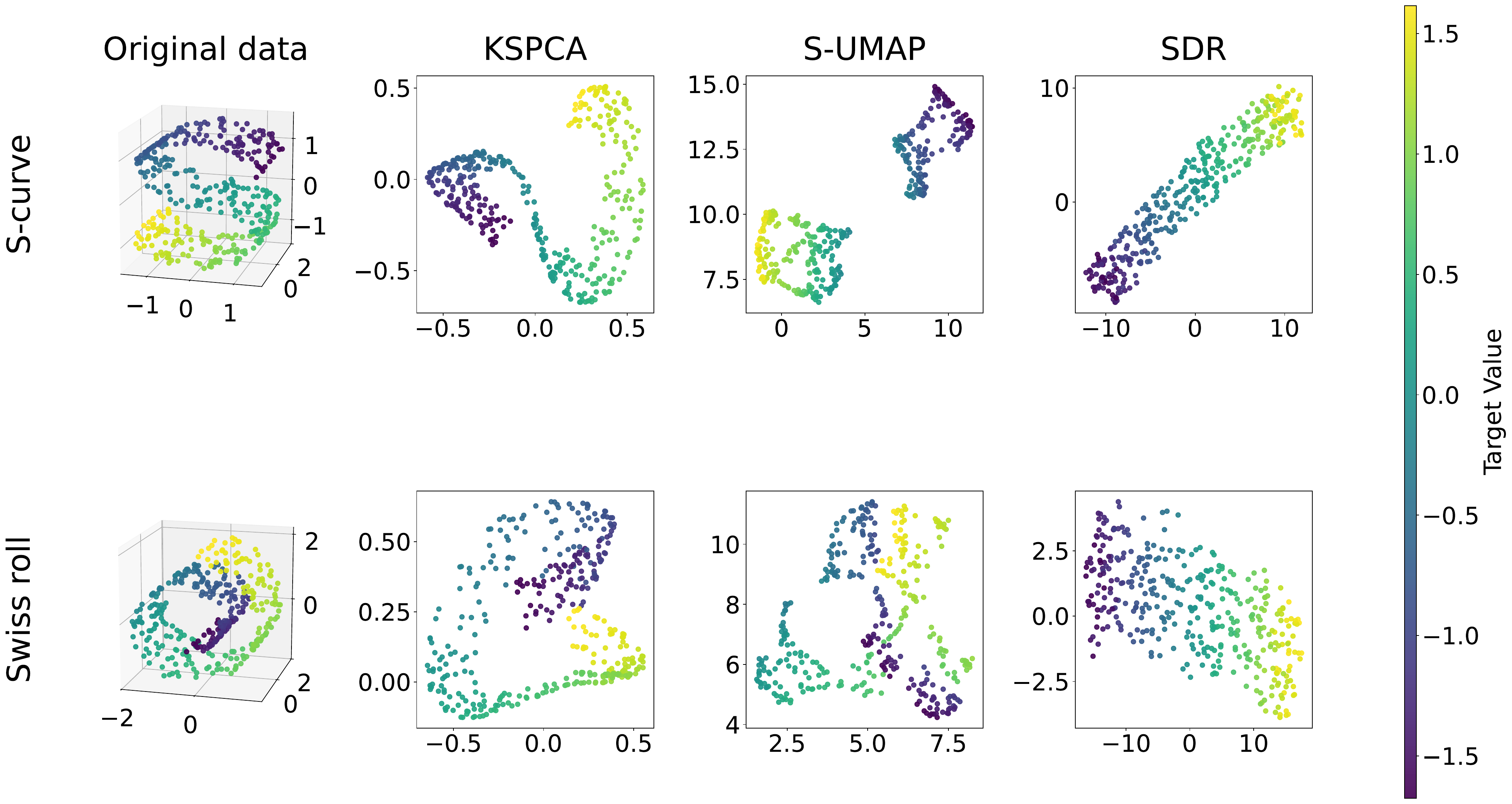}
    \caption{Visualizing embeddings in the pure DR setting for KSPCA, (S-UMAP) and SDR on toy 3D datasets.}  
    \label{fig:toy_data}
\end{figure}

SDR just like DistR inherits the same runtime complexity $\mathcal{O}(n^{2}m + nm^{2})$ where $n$ is the number of source samples and $m$ is the number of target samples. For the pure DR case ($p < d, n = m$), SDR scales as $\mathcal{O}(n^{3})$. This is one of the core reasons why SDR is feasible as a distributional reduction method but not very scalable as a DR method. However, SDR might still be a very useful DR method in the low-data regime. While this is evident from the experiments in Section \ref{subsec:kl_expts}, we show visualizations of two toy datasets namely: (i) S-curve and (ii) Swiss Roll datasets to show how SDR behaves as a DR method when compared with other supervised DR methods in Figure \ref{fig:toy_data}. We chose these datasets, since they are the most basic datasets tested for every DR method and is easier to visualize the embeddings along with the original data since the original data lives in 3D. We compare the embeddings generated from Supervised-UMAP (S-UMAP) and Kernel Supervised PCA (KS-PCA) visually. For the S-curve dataset, we set the number of neighbours to 50 and for the swiss roll dataset we set this parameter to 20 to generate the supervised UMAP visualizations.

 For SDR, we choose the exact same settings used in the kernel learning experiments. The expected behaviour for any DR method on these two datasets is to unroll the curves onto flat rectangular strips in the 2D space. We can see that SDR achieves this quite cleanly compared to S-UMAP or KS-PCA. Along with this we also use the Friedman dataset \citep{friedman1991dataset} for testing models on a downstream predictive task. We test the predictive performance of a k-Nearest Neighbours (k-NN) model and a Kernel Ridge Regression (KRR) model to show that the SDR embeddings work well in a model-agnostic manner. We report the $R^{2}$ scores for this experiment in Tables \ref{tab:knn_results} and \ref{tab:krr_results} and the metrics reported are across 5 different random seeds.

\begin{table}[h!]
\captionsetup{skip=12pt}
\centering
\resizebox{\textwidth}{!}{
\begin{tabular}{lcc cc cc}
\toprule
 & \multicolumn{2}{c}{S-curve (100 points)} & \multicolumn{2}{c}{Swiss Roll (100 points)} & \multicolumn{2}{c}{Friedman (500 points)} \\
\cmidrule(lr){2-3} \cmidrule(lr){4-5} \cmidrule(lr){6-7}
 & Train & Test 
 & Train & Test 
 & Train & Test \\
\midrule
Supervised UMAP & \textbf{0.99} $\pm$ \textbf{0.00} & 0.88 $\pm$ 0.08 & 0.99 $\pm$ 0.00 & 0.47 $\pm$ 0.49 & \textbf{0.99} $\pm$ \textbf{0.00} & 0.51 $\pm$ 0.14 \\
KS-PCA & 0.92 $\pm$ 0.06 & 0.93 $\pm$ 0.05 & 0.89 $\pm$ 0.10 & 0.85 $\pm$ 0.10 & 0.81 $\pm$ 0.02 & 0.71 $\pm$ 0.10 \\
SDR & 0.97 $\pm$ 0.02 & \textbf{0.94} $\pm$ \textbf{0.06}& 0.99 $\pm$ 0.01 & \textbf{0.97} $\pm$ \textbf{0.03} & 0.98 $\pm$ 0.00 & \textbf{0.91} $\pm$ \textbf{0.03} \\
\bottomrule
\end{tabular}}
\caption{$R^{2}$ scores for k-NN Regressor with $k = 10$.}
\label{tab:knn_results}
\end{table}

\begin{table}[h!]
\captionsetup{skip=12pt}
\centering
\resizebox{\textwidth}{!}{
\begin{tabular}{lcc cc cc}
\toprule
 & \multicolumn{2}{c}{S-curve (100 points)} & \multicolumn{2}{c}{Swiss Roll (100 points)} & \multicolumn{2}{c}{Friedman (500 points)} \\
\cmidrule(lr){2-3} \cmidrule(lr){4-5} \cmidrule(lr){6-7}
 & Train & Test 
 & Train & Test 
 & Train & Test \\
\midrule
Supervised UMAP & 0.97 $\pm$ 0.01 & 0.82 $\pm$ 0.13 & 0.98 $\pm$ 0.01 & 0.53 $\pm$ 0.34 & 0.98 $\pm$ 0.00 & 0.55 $\pm$ 0.12 \\
KS-PCA & 0.90 $\pm$ 0.06 & 0.92 $\pm$ 0.06 & 0.86 $\pm$ 0.12 & 0.82 $\pm$ 0.12 & 0.80 $\pm$ 0.03 & 0.72 $\pm$ 0.08 \\
SDR & \textbf{0.98} $\pm$ \textbf{0.01} & \textbf{0.96} $\pm$ \textbf{0.03} & \textbf{0.99} $\pm$ \textbf{0.01} & \textbf{0.96} $\pm$ \textbf{0.05} & 0.98 $\pm$ 0.00 & \textbf{0.91} $\pm$ \textbf{0.02} \\
\bottomrule
\end{tabular}}
\caption{$R^{2}$ scores for KRR Regressor with $\alpha =1 $ (lengthscale of RBF kernel).}
\label{tab:krr_results}
\end{table}

In several cases, SDR is better than Supervised UMAP on the test $R^{2}$ scores and is comparable on the train $R^{2}$ scores. This also shows that the mechanism of optimizing for $L$ not only provides a mechanism to project OOS points onto the embedding space but also regularizes the optimized embeddings $Z$ resulting in better test performance for a downstream predictor.

\section{Kernel Learning Experimental Details}
\label{appx:kl_expts}
\subsection{Datasets}
The experiments comparing SDR-GP with all the other methods included 5 datasets where 3 of them are regression datasets sourced from the UCI repository and 2 of them are classification datasets. The details of these datasets are provided in Tables \ref{tab:reg_datasets} and \ref{tab:classify_datasets}.

\begin{table}[h!]
\captionsetup{skip=10pt}
    \centering
    \begin{tabular}{|c|c|c|c|}
        \hline
         \textbf{Dataset} & \textbf{Number of Points} & \textbf{Input Features} & \textbf{Output Features} \\
         \hline
         Boston Housing & 506 & 13 & 1 \\
         \hline
         Energy Efficiency & 768 & 8 & 2 \\
         \hline
         Concrete Compressive Strength & 1030 & 8 & 1 \\
         \hline
    \end{tabular}
    \caption{Details of Regression Datasets used in Kernel Learning experiments}
    \label{tab:reg_datasets}
\end{table}
\begin{table}[h!]
\captionsetup{skip=10pt}
    \centering
    \begin{tabular}{|c|c|c|c|}
        \hline
         \textbf{Dataset} & \textbf{Number of Points} & \textbf{Input Features} & \textbf{Number of Classes} \\
         \hline
         MNIST & 2000 & 784 (28$\times$28) & 10 \\
         \hline
         COIL-20 & 1440 & 16384 (128$\times$128) & 20 \\
         \hline
    \end{tabular}
    \caption{Details of Classification (Image) Datasets used in Kernel Learning experiments}
    \label{tab:classify_datasets}
\end{table}

\subsection{Models}
We compare 5 different models in this experiment. We compare SDR-GP against Deep Kernel Learning \citep{wilson16dkl} and Deep Gaussian Processes (\cite{damianou2013deepgps}, \cite{salimbeni2017dgpvi}) which are the most natural ways to add non-stationarity to GPs for modelling non-smooth functions. Since SDR also acts as a pure DR method in this application, we compare against two other supervised DR methods namely Kernel Supervised PCA \citep{barshan2011spca} and Supervised UMAP \citep{mcinnes2020umap} that act as embedding models/feature extractors for the GP. We dub these models KSPCA-GP and UMAP-GP respectively. For KSPCA, we use RBF kernels for the inputs, RBF and Delta kernels on the targets for regression and classification datasets respectively. For supervised UMAP, we use the official UMAP\footnote[4]{https://umap-learn.readthedocs.io/en/latest/index.html} library. We set the number of neighbours to 50, the effective minimum distance between embedded points to 0.1 and the distance metric to 'L2'.

\subsection{NN architectures and GP settings}
For regression tasks, we use the same NN Feature Extractor which is a Multilayer-Perceptron network with ReLU activations used in \cite{wilson16dkl}. The network is a fully connected DNN with [d-1000-500-50-2] architecture. The GP for the Boston and the Concrete datasets are exact GPs that use an RBF kernel with Automatic Relevance Determination (ARD) \citep{Rasmussen2006Gaussian} and the standard Gaussian likelihood. The GP for the Energy dataset is a multi-task GP \citep{bonilla2007mtgp} that models both the covariances between the inputs and the output tasks with an RBF kernel (equipped with ARD) and a Gaussian likelihood.

The Deep GP model was constructed with one hidden layer and an output layer. The latent dimensions of the hidden layer were set to match the latent dimensions of all the other models. This model is trained variationally using the doubly stochastic variational inference technique described by \cite{salimbeni2017dgpvi}. The number of inducing points is set to 128 and the number of likelihood samples is set to 16.

For the classification tasks, we again use the same CNN feature extractor described in \citep{wilson16dkl} (Appendix A.1 in the paper). However, we use this feature extractor for classification instead of orientation extraction as described in the DKL paper. We use an exact GP with a Dirichlet likelihood for multi-class classification inspired by the work of \cite{milios2018dirichletgp}. We set $\alpha_{\epsilon}$ (concentration parameter of the Dirichlet likelihood) to 0.01 and the number of samples from the GP posterior to 256 since the posterior predictions cannot be obtained in closed form and needs to be MC-sampled.

\subsection{SDR settings}
For SDR, we use the Symmetric Entropic Affinity(SEA) from \cite{vanassel2023sea} on the inputs $\textbf{X}$ for all experiments in this section with the perplexity and the learning rate tuned for each dataset. The perplexity value was set to 20 for all experiments. The learning rate parameter was set to 0.01 for the regression datasets and 0.1 for the image datasets. The normalized Student-t affinity was used to compute affinities for the embeddings $\textbf{Z}$. For the HSIC computation, we use the Rational Quadratic (RQ) kernel for the embedding space and the RBF/Delta kernel \citep{barshan2011spca} for the regression/classification outputs. The kernel $\textbf{K}$ in the SDR-OOS objective is computed with RBF for all the experiments. The $\alpha$ parameter in the srBSFGW objective is set to 0.2, $\eta$ is 
set to 1000, $\beta$ is set to 0.5 and $\lambda_{L}$ was set to $1e-2$ for all experiments.
The coupling matrix $\textbf{T}$ is initialized using spectral clustering for regression tasks.
$\textbf{T}$ is initialized randomly for classification tasks since any clustering technique can run into the issue of producing degenerate clusters when the number of clusters required is much higher than the number of actual clusters present in the data which is typically the case with classification and the pure DR setting.

\subsection{Experimental Details}
 For the regression datasets, the number of embedding dimensions was set to 2 and for the classification datasets the number of embedding dimensions was set to 10 following the practice in \cite{ober2021pitfallsdkl}. All the GP models were trained for 20 epochs for the regression benchmarks and were trained for 50 epochs for the classification benchmarks. The learning rate was set to 0.1 for all the models. The test data size was set to 0.2 for all experiments. We were unable to obtain stable DGP results within our compute budget for the classification setting due to memory pressure when scaling the number of latent dimensions to 10. This experiment was run on a remote server equipped with an NVIDIA A100-80GB GPU so that the Dirichlet GP fits in memory for the image classification datasets.

\subsection{Model training times}
We report the training times for the various models that were considered for the kernel learning benchmarks (Section \ref{subsec:kl_expts}) in Table \ref{tab:kl_runtimes}. The training times reported for KSPCA-GP, UMAP-GP and SDR-GP includes the time taken to fit the DR transform and the time taken to train the GP since the fitting of the DR transform is decoupled from GP training in these models.

\begin{table}[h!]
\captionsetup{skip=12pt}
\centering
\resizebox{\textwidth}{!}{
\begin{tabular}{lccccc}
\toprule
 & Boston & Energy & Concrete & COIL-20 & MNIST-2k \\
\midrule
NCA-GP & 0.45 $\pm$ 0.07 & 1.34 $\pm$ 0.30 & 3.00 $\pm$ 1.20 & 1.76 $\pm$ 0.13 & 2.62 $\pm$ 0.24 \\
KSPCA-GP & 0.15 $\pm$ 0.08 & 0.61 $\pm$ 0.01 & 0.49 $\pm$ 0.05 & 1.27 $\pm$ 0.04 & 1.09 $\pm$ 0.03 \\
UMAP-GP  & 8.01 $\pm$ 16.42 & 1.70 $\pm$ 0.05 & 1.93 $\pm$ 0.15 & 16.59 $\pm$ 12.10 & 6.80 $\pm$ 0.10 \\
DGP & 0.44 $\pm$ 0.07 & 0.80 $\pm$ 0.01 & 0.41 $\pm$ 0.00 & - & - \\
DKL & 0.53 $\pm$ 0.89 & 0.56 $\pm$ 0.02 & 0.39 $\pm$ 0.03 & 34.63 $\pm$ 0.31 & 2.40 $\pm$ 0.06 \\
SDR-GP (ours) & 7.07 $\pm$ 2.21 & 5.14 $\pm$ 0.60 & 8.42 $\pm$ 2.44 & 8.00 $\pm$ 0.57 & 17.47 $\pm$ 2.55 \\
\bottomrule
\end{tabular}}
\caption{Runtime (in seconds) for all datasets based on five random seeds}
\label{tab:kl_runtimes}
\end{table}

\section{Uncertainty Calibration Diagnostics}
\label{appx:calibration}
\begin{figure}[h!]
    \centering
    \includegraphics[width=0.89\linewidth]{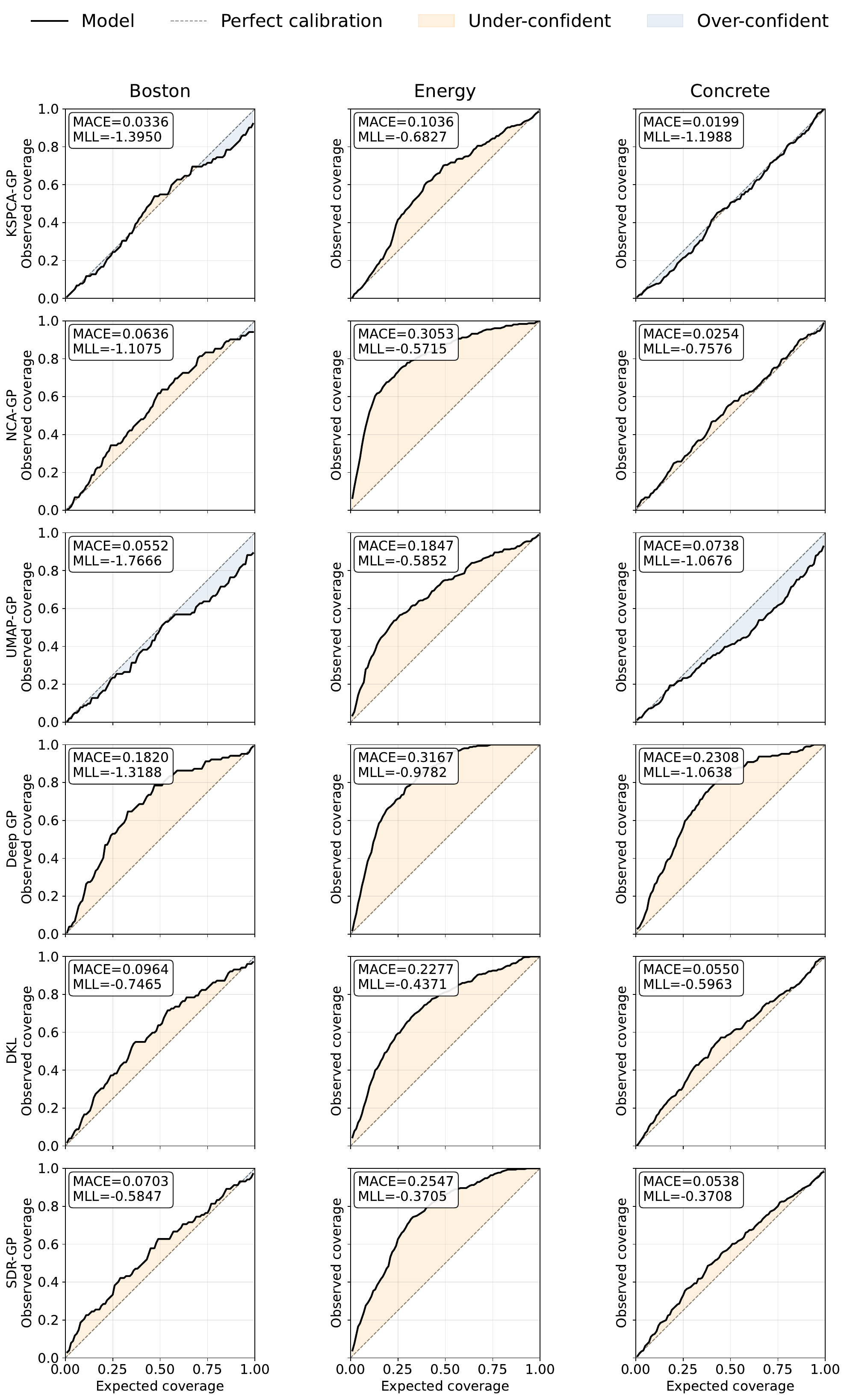}
    \caption{Calibration curves for the UCI regression datasets. The dotted diagonal  line is the ideal calibration curve. Anything above the diagonal means the model is under-confident and anything below the diagonal means the model is over-confident.}
    \label{fig:calibration_curves}
\end{figure}

When assessing probabilistic ML models, uncertainty calibration plays a major role in analyzing the predictive uncertainties produced by a model. For this, we use standard calibration curves and Mean Absolute Calibration Error as the metric. Figure \ref{fig:calibration_curves} shows typical calibration curves of all the models in our experimental suite for the UCI regression datasets.

A calibration curve fundamentally validates the fact: \textbf{"An $\alpha \%$ prediction interval should contain $\alpha \%$" of the true values"}. For a scalar metric, we compute MACE following the uncertainty-toolbox convention of \cite{chung2021uncertaintytoolbox}, based on the regression calibration framework of \cite{kuleshov18uq}. For Gaussian predictive uncertainties, this corresponds to comparing nominal Gaussian interval/quantile probabilities with empirical coverage, as in \cite{tran2020uqcomparison}. In this work, we stick to the symmetric interval calibration mode for generating the calibration curves. The number of levels across which we assess the calibration is set to $n_{bins} = 99$. For $n_{bins} = 99$, we consider uniformly spaced confidence intervals $\alpha_{k} \in (0, 1)$. The "Interval" mode constructs symmetric prediction bands as follows:
\begin{equation}
    [\hat{y} - z_{\alpha_{k}}\hat{\sigma}, \hat{y} + z_{\alpha_{k}}\hat{\sigma}], \ z_{\alpha_{k}} = \Phi^{-1}\left(\frac{1 + \alpha_{k}}{2}\right)
\end{equation}
where $\hat{y}$ is the predicted mean, $\hat{\sigma}$ is the predicted standard deviation and $\Phi^{-1}$ is the inverse CDF of a Gaussian distribution. 

Figure~\ref{fig:calibration_curves} reports observed against expected coverage for each embedding--GP pipeline. SDR-GP attains the highest mean log-likelihood on all three datasets, improving on the strongest competing pipeline, DKL, while its miscalibration area remains comparable to the other deep baselines and uniformly below that of Deep~GP. Its coverage curves lie above the diagonal throughout, so the residual miscalibration is under-confidence: the predictive intervals are conservative rather than optimistic, which is the preferable failure mode in a downstream decision setting, and SDR-GP never exhibits the over-confident regions visible for KSPCA-GP and UMAP-GP on Boston and Concrete. We note that MACE assesses marginal coverage only and is insensitive to sharpness; KSPCA-GP illustrates the distinction, attaining the lowest MACE on every dataset while yielding the least informative predictive densities. Energy is the hardest case for all methods, with every pipeline markedly under-confident.

\section{Comparison with pre-trained feature extractors for DKL}
\label{appx:pretrained_dkl}
A widely used variant of DKL is where the feature extractor networks are pre-trained and then fine-tuned jointly along with the GP hyperparameters instead of training them from scratch. We use the same networks used for the experiments in Section 5.2 but pre-train them using squared loss and cross-entropy loss for the regression and classification benchmarks respectively. We train them during the pre-training phase for 75 epochs and with a learning rate of $1e-3$ followed by a training phase of 20 epochs for the regression datasets and 50 epochs for the classification datasets with a learning rate of 0.01. The results are show in Tables \ref{tab:dkl_variants_reg} and \ref{tab:dkl_variants_classify} where DKL-joint corresponds to the model where the NN feature extractor is trained from scratch jointly with the GP hyperparameters and DKL-pretrained corresponds to the model where the NN feature extractor is pre-trained and then fine-tuned along with the GP hyperparameters in the joint-training phase.

\begin{table}[h!]
\captionsetup{skip=12pt}
\centering
\resizebox{\textwidth}{!}{
\begin{tabular}{lcc cc cc}
\toprule
 & \multicolumn{2}{c}{Boston} & \multicolumn{2}{c}{Energy} & \multicolumn{2}{c}{Concrete} \\
\cmidrule(lr){2-3} \cmidrule(lr){4-5} \cmidrule(lr){6-7}
 & MLL ($\uparrow$) & MSE ($\downarrow$) 
 & MLL ($\uparrow$) & MSE ($\downarrow$) 
 & MLL ($\uparrow$) & MSE ($\downarrow$) \\
\midrule
DKL-joint & -0.42 $\pm$ 0.21 & 0.13 $\pm$ 0.07 & -0.45 $\pm$ 0.03 & 0.09 $\pm$ 0.02 & -0.65 $\pm$ 0.11 & 0.21 $\pm$ 0.05 \\
DKL-pretrained & -0.76 $\pm$ 0.05 & 0.11 $\pm$ 0.07 & -0.71 $\pm$ 0.10 & 0.08 $\pm$ 0.02 & -0.82 $\pm$ 0.01 & 0.17 $\pm$ 0.01 \\
SDR-GP (ours) & -0.32 $\pm$ 0.18 & 0.14 $\pm$ 0.08 & -0.37 $\pm$ 0.02 & 0.05 $\pm$ 0.01 & -0.40 $\pm$ 0.07 & 0.13 $\pm$ 0.02 \\
\bottomrule
\end{tabular}}
\caption{Test results for DKL variants on UCI regression datasets based on five random seeds}
\label{tab:dkl_variants_reg}
\end{table}

\begin{table}[h!]
\captionsetup{skip=12pt}
\centering
\small
{\setlength{\tabcolsep}{4pt}
\begin{tabular}{lcccc}
\toprule
 & \multicolumn{2}{c}{MNIST-2k} & \multicolumn{2}{c}{COIL-20} \\
\cmidrule(lr){2-3} \cmidrule(lr){4-5}
 & MLP ($\uparrow$) & ACC ($\uparrow$) 
 & MLP ($\uparrow$) & ACC ($\uparrow$) \\
\midrule
DKL-joint & -0.30 $\pm$ 0.05 & 0.90 $\pm$ 0.02 & -0.09 $\pm$ 0.02 & 0.99 $\pm$ 0.00 \\
DKL-pretrained & -0.25 $\pm$ 0.10 & 0.94 $\pm$ 0.03 & -0.13 $\pm$ 0.01 & 0.97 $\pm$ 0.03 \\
SDR-GP (ours) & -0.23 $\pm$ 0.02 & 0.95 $\pm$ 0.00 & -0.07 $\pm$ 0.02 & 0.99 $\pm$ 0.00 \\
\bottomrule
\end{tabular}
}
\caption{Test results for DKL variants on image classification datasets based on five random seeds.}
\label{tab:dkl_variants_classify}
\end{table}

These results emphasise that the performance of improvements of SDR are not solely attributable to the joint feature learning regime of DKL. Even in this controlled setting where both SDR and DKL operate on comparable pretrained features, SDR consistently achieves comparable or better predictive performances. These results suggest that the observed gains arise from the proposed SDR formulation itself rather than differences in representation learning or end-to-end optimization dynamics. 

\section{Code/Implementation details}
\label{appx:code}
The implementation of SDR was built on top of the DistR\footnote[1]{https://github.com/huguesva/Distributional-Reduction} implementation from \cite{vanassel2025distr}. The benchmarking scripts for kernel learning use GP models constructed with GPyTorch\footnote[2]{https://docs.gpytorch.ai/en/stable/}\citep{gardner2018gpytorch}. For solving the srBSFGW objective, we use the FGW/semi-relaxed FGW solvers from the Python Optimal Transport (POT)\footnote[3]{https://pythonot.github.io} library. All experiments were implemented in Python using PyTorch for automatic differentiation and GPU acceleration. Experiments were run on NVIDIA GPUs; the distributional reduction benchmarks used an NVIDIA L4 GPU, while the kernel learning benchmarks used an NVIDIA A100-80GB GPU. All reported means and standard deviations (1-sigma error bars) are computed over multiple random seeds, with the number of seeds specified in the corresponding experimental details.

The code for the SDR implementation and benchmarking scripts is available at: \href{https://github.com/SaiAakash/supervised-distributional-reduction}{https://github.com/SaiAakash/supervised-distributional-reduction}.


\newpage
\section*{NeurIPS Paper Checklist}

\begin{enumerate}

\item {\bf Claims}
    \item[] Question: Do the main claims made in the abstract and introduction accurately reflect the paper's contributions and scope?
    \item[] Answer: \answerYes{}
    \item[] Justification: The abstract and introduction describe the SDR method, and an out-of-sample extension for SDR with application to GPs. Sections \ref{sec:Method} and \ref{sec:oos-extension} talk about these key contributions in detail. Experiments in Section \ref{sec:expts} show results on these main claims with the proof for Proposition \ref{prop:proto_targets} in Appendix \ref{appx:prop1}.

\item {\bf Limitations}
    \item[] Question: Does the paper discuss the limitations of the work performed by the authors?
    \item[] Answer: \answerYes{}
    \item[] Justification: Section \ref{subsec:kl_expts} and Appendix \ref{appx:dr_results} talk about the scalability of SDR as a pure DR method. Section \ref{sec:conclusion} also proposes as future work, potential avenues to reduce runtime complexity and provide convergence guarantees for this framework.

\item {\bf Theory assumptions and proofs}
    \item[] Question: For each theoretical result, does the paper provide the full set of assumptions and a complete (and correct) proof?
    \item[] Answer: \answerYes{}
    \item[] Justification: The paper contains the proof for Proposition \ref{prop:proto_targets} in Appendix \ref{appx:prop1}. All the underlying assumptions for this Proposition are clearly specified in the corresponding statement.

    \item {\bf Experimental result reproducibility}
    \item[] Question: Does the paper fully disclose all the information needed to reproduce the main experimental results of the paper to the extent that it affects the main claims and/or conclusions of the paper (regardless of whether the code and data are provided or not)?
    \item[] Answer: \answerYes{}
    \item[] Justification: Appendix \ref{appx:code} provides basic details of the code implementation for the method and benchmarks. Appendix \ref{appx:distr_expts} and \ref{appx:kl_expts} provide detailed information of the datasets used and all the settings required to reproduce the results of the Distributional Reduction and Kernel Learning benchmarks shown in the paper.

\item {\bf Open access to data and code}
    \item[] Question: Does the paper provide open access to the data and code, with sufficient instructions to faithfully reproduce the main experimental results, as described in supplemental material?
    \item[] Answer: \answerYes{}
    \item[] Justification: A link to the code repository has been provided in Appendix \ref{appx:code}. The repository contains all the datasets used for the experiments with clear instructions to run the benchmarking scripts.

\item {\bf Experimental setting/details}
    \item[] Question: Does the paper specify all the training and test details (e.g., data splits, hyperparameters, how they were chosen, type of optimizer) necessary to understand the results?
    \item[] Answer: \answerYes{}
    \item[] Justification: While Section \ref{subsec:distr_benchmarks} and Section \ref{subsec:kl_expts} explain the experimental setting and the evaluation metrics briefly, Appendix \ref{appx:distr_expts} and \ref{appx:kl_expts} explain in detail the experimental setting, evaluation metrics and the complete settings for the SDR algorithm, GP models and the baseline methods to understand the results in the main paper. 

\item {\bf Experiment statistical significance}
    \item[] Question: Does the paper report error bars suitably and correctly defined or other appropriate information about the statistical significance of the experiments?
    \item[] Answer: \answerYes{}
    \item[] Justification: The paper reports mean and standard deviations (1-sigma error bars) for all the evaluation metrics in the various experiments across different random seeds. We also plot the uncertainty bands in the various graphs to indicate the same.

\item {\bf Experiments compute resources}
    \item[] Question: For each experiment, does the paper provide sufficient information on the computer resources (type of compute workers, memory, time of execution) needed to reproduce the experiments?
    \item[] Answer: \answerYes{}
    \item[] Justification: Appendix \ref{appx:code} talks about the compute resources used to run the experiments (these are also briefly mentioned in Appendix \ref{appx:distr_expts} and \ref{appx:kl_expts}).
    
\item {\bf Code of ethics}
    \item[] Question: Does the research conducted in the paper conform, in every respect, with the NeurIPS Code of Ethics \url{https://neurips.cc/public/EthicsGuidelines}?
    \item[] Answer: \answerYes{}
    \item[] Justification: The research involves algorithmic development and evaluation on standard regression and classification benchmarks. It does not involve human subjects or obviously ethically sensitive applications, and we assume it conforms to the NeurIPS Code of Ethics.

\item {\bf Broader impacts}
    \item[] Question: Does the paper discuss both potential positive societal impacts and negative societal impacts of the work performed?
    \item[] Answer: \answerNo{}
    \item[] Justification: This work focuses on a theoretical and methodological contribution to distributional reduction, without a specific application domain or deployment context. As such, we do not identify direct or immediate positive or negative societal impacts; any downstream effects would depend on how the method is applied in practice, which is beyond the scope of this paper.
    
\item {\bf Safeguards}
    \item[] Question: Does the paper describe safeguards that have been put in place for responsible release of data or models that have a high risk for misuse (e.g., pre-trained language models, image generators, or scraped datasets)?
    \item[] Answer: \answerNA{}
    \item[] Justification: The paper does not involve any foundational or pre-trained models that are more widely usable and hence, safeguards are not applicable for this paper.

\item {\bf Licenses for existing assets}
    \item[] Question: Are the creators or original owners of assets (e.g., code, data, models), used in the paper, properly credited and are the license and terms of use explicitly mentioned and properly respected?
    \item[] Answer: \answerYes{}
    \item[] Justification: Appendix \ref{appx:code} cites the previous code base on top of which the implementation of the current work was built. Appendix \ref{appx:distr_expts} and \ref{appx:kl_expts} cite the corresponding work for all of the datasets used in this work. The license and terms of these datasets are attached in the supplementary code as well. 

\item {\bf New assets}
    \item[] Question: Are new assets introduced in the paper well documented and is the documentation provided alongside the assets?
    \item[] Answer: \answerNA{}
    \item[] Justification: There are no new assets introduced in this paper and therefore this is not applicable for this work.

\item {\bf Crowdsourcing and research with human subjects}
    \item[] Question: For crowdsourcing experiments and research with human subjects, does the paper include the full text of instructions given to participants and screenshots, if applicable, as well as details about compensation (if any)? 
    \item[] Answer: \answerNA{}.
    \item[] Justification: The paper does not involve any experiments requiring human subjects and so this requirement is not applicable.

\item {\bf Institutional review board (IRB) approvals or equivalent for research with human subjects}
    \item[] Question: Does the paper describe potential risks incurred by study participants, whether such risks were disclosed to the subjects, and whether Institutional Review Board (IRB) approvals (or an equivalent approval/review based on the requirements of your country or institution) were obtained?
    \item[] Answer:  \answerNA{}.
    \item[] Justification: The research does not involve human subjects, therefore IRB approval is not applicable.

\item {\bf Declaration of LLM usage}
    \item[] Question: Does the paper describe the usage of LLMs if it is an important, original, or non-standard component of the core methods in this research? Note that if the LLM is used only for writing, editing, or formatting purposes and does \emph{not} impact the core methodology, scientific rigor, or originality of the research, declaration is not required.
    \item[] Answer: \answerYes{}
    \item[] Justification: Apart from using LLMs for editing and formatting parts of the paper, they were also used to create some visualizations and facilitate the running of experiments.

\end{enumerate}

\end{document}